\documentclass[10pt,twocolumn,letterpaper]{article}

\usepackage{iccv}
\usepackage{times}
\usepackage{epsfig}
\usepackage{graphicx}
\usepackage{amsmath}
\usepackage{amssymb}

\usepackage[utf8]{inputenc}
\usepackage{multirow}
\usepackage[caption=false]{subfig}
\usepackage{verbatim}
\usepackage{color}
\usepackage{wrapfig}
\usepackage{booktabs}
\usepackage{array}
\usepackage{diagbox}
\usepackage[T1]{fontenc}    
\usepackage{url}            
\usepackage{booktabs}       
\usepackage{amsfonts}       
\usepackage{nicefrac}       
\usepackage{microtype}      
\usepackage{setspace}
\usepackage{multirow}
\usepackage{amssymb}
\usepackage{diagbox}
\usepackage{subfig}
\usepackage{wrapfig}
\usepackage{diagbox} 
\usepackage{wrapfig,lipsum,booktabs}
\usepackage{caption}
\usepackage{colortbl}

\usepackage{color}
\usepackage{xcolor}
\definecolor{citecolor}{HTML}{0071bc}

\usepackage[pagebackref=true,breaklinks=true,colorlinks,citecolor=citecolor,bookmarks=false]{hyperref}

\definecolor{demphcolor}{RGB}{144,144,144}
\definecolor{obmanblue}{RGB}{145,191,219}
\definecolor{hored}{RGB}{240,128,128}
\definecolor{fphapurple}{RGB}{190,150,220}
\definecolor{tred}{RGB}{250,100,100}

\newcommand{\demph}[1]{\textcolor{demphcolor}{#1}}

\newcolumntype{x}[1]{>{\centering\arraybackslash}p{#1pt}}

\newcommand{\app}{\raise.17ex\hbox{$\scriptstyle\sim$}}

\newlength\savewidth\newcommand\shline{\noalign{\global\savewidth\arrayrulewidth
  \global\arrayrulewidth 1pt}\hline\noalign{\global\arrayrulewidth\savewidth}}
\newcommand{\tablestyle}[2]{\setlength{\tabcolsep}{#1}\renewcommand{\arraystretch}{#2}\centering\footnotesize}

\usepackage{amssymb}
\usepackage{pifont}

\newcommand{\customfootnotetext}[2]{{
  \renewcommand{\thefootnote}{#1}
  \footnotetext[0]{#2}}}
  
\iccvfinalcopy 

\def\iccvPaperID{4376} 

\ificcvfinal\fi

\begin{document}

\title{Hand-Object Contact Consistency Reasoning for Human Grasps Generation}

\author{
Hanwen Jiang\textsuperscript{*} \quad
Shaowei Liu\textsuperscript{*} \quad 
Jiashun Wang \quad
Xiaolong Wang \\
UC San Diego
}

\twocolumn[{%
\maketitle

\begin{center}
    \includegraphics[width=\linewidth]{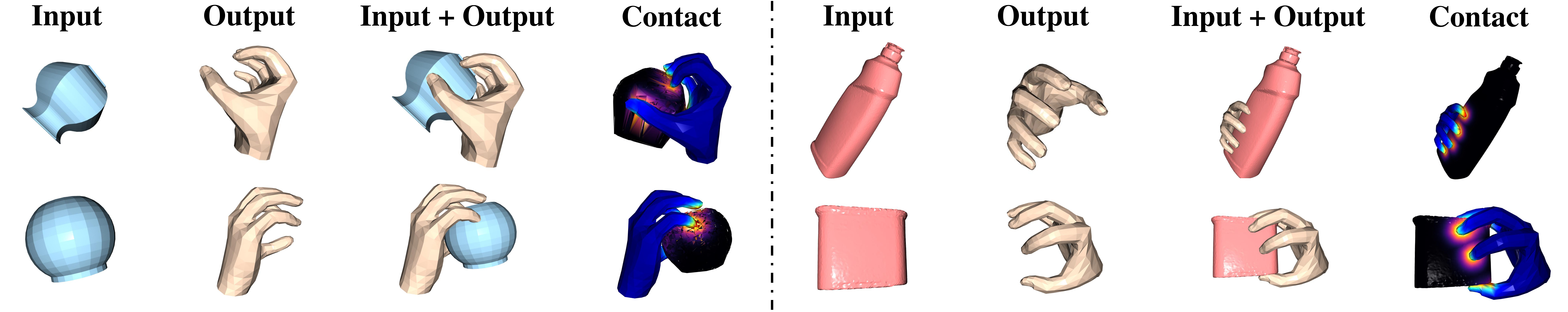}
    \vspace{-0.2in}
    \captionof{figure}{\small{Generated human grasp on \textcolor{obmanblue}{in-domain} and \textcolor{hored}{out-of-domain} objects. 
    Object and hand contact maps are shown in the last column. The brighter the region is, the higher the contact values between the hand and object are. Best viewed in color.}}
    \label{fig:teaser}
\end{center}
}]
\customfootnotetext{*}{Equal contribution.}

\maketitle
\ificcvfinal\fi

\begin{abstract}
\vspace{-0.05in}
While predicting robot grasps with parallel jaw grippers have been well studied and widely applied in robot manipulation tasks, the study on natural human grasp generation with a multi-finger hand remains a very challenging problem. In this paper, we propose to generate human grasps given a 3D object in the world. Our key observation is that it is crucial to model the consistency between the hand contact points and object contact regions. That is, we encourage the prior hand contact points to be close to the object surface and the object common contact regions to be touched by the hand at the same time. Based on the hand-object contact consistency, we design novel objectives in training the human grasp generation model and also a new self-supervised task which allows the grasp generation network to be adjusted even during test time. Our experiments show significant improvement in human grasp generation over state-of-the-art approaches by a large margin. More interestingly, by optimizing the model during test time with the self-supervised task, it helps achieve larger gain on unseen and out-of-domain objects. Project page: \href{https://hwjiang1510.github.io/GraspTTA/}{https://hwjiang1510.github.io/GraspTTA/}.
\end{abstract}

\section{Introduction}
\vspace{-0.05in}
Capturing hand-object interactions has been an active field of study~\cite{tekin2019h+, hasson2019learning, liu2021semi, Doosti2020HOPENetAG, brahmbhatt2019contactdb, brahmbhatt2020contactpose, taheri2020grab, Schrder2017HandObjectID, Yang2015GraspTR} and it has wide applications in virtual reality~\cite{Hll2018EfficientPI, Wu2020HandPE}, human-computer interaction~\cite{Ueda2003AHE} and imitation learning in robotics~\cite{Zhang2018DeepIL, Thobbi2010ImitationLO, Radosavovic2020StateOnlyIL}. In this paper, we study the interactions via generation: As shown in Fig.~\ref{fig:teaser}, given only a 3D object in the world coordinate, we generate the 3D human hand for grasping it. Unlike predicting robot grasps with parallel jaw grippers~\cite{mousavian20196, Yan2018Learning6G, zhou20176dof, cao2021suctionnet1billion}, predicting human grasps is substantially more difficult because: (i) Human hands have a lot more degrees of freedom, which leads to much more complex contact; (ii) The generated grasp needs to be not only physically plausible but also presented in a natural way, consistent with how objects are usually grasped. 

To synthesize physically plausible and natural grasp poses, recent works propose to use generative models~\cite{corona2020ganhand,karunratanakul2020grasping, taheri2020grab} supervised by large-scale datasets~\cite{hasson2019learning, Hampali2019HOnnotateAM, GarciaHernando2018FirstPersonHA} with grasp annotations and contact analysis on hands. Specifically, the large-scale dataset allows the model to generate realistic human grasps and the contact analysis encourages the hand contact points to be close with the object but without inter-penetration. While these methods put a lot of efforts into modeling the hand and its contact points, they ignore that the object itself also has more possible contact regions that need to be reached (see contact map in Fig.~\ref{fig:teaser}). In fact, recent work has studied the common contact regions on objects and trained neural networks to directly predict the contact map from the 3D object model~\cite{brahmbhatt2019contactdb, brahmbhatt2020contactpose}. 



In this paper, we argue that it is critical for the hand contact points and object contact regions to reach mutual agreement and consistency for grasp generation. To achieve this, we propose to unify two separate models for both the hand grasp synthesis and object contact map estimation. We show that the consistency constraint between hand contact points and object contact map is not only useful for optimizing better grasps during training time by designing new losses, but also provides a self-supervised task to adjust the grasp when testing on a novel object. We introduce the two components as follows. 

First, we train a Conditional Variational Auto-Encoder~\cite{Sohn2015LearningSO} (CVAE) based network which takes the 3D object point clouds as inputs and predicts the hand grasp parameterized by a MANO model~\cite{romero2017embodied}, namely GraspCVAE. During training the GraspCVAE, we design two novel losses with one encouraging the hand to touch the object surface and another forcing the object contact regions touched by the ground truth hand close to the predicted hand. With these two consistent losses, we observe more realistic and physically plausible grasps.

Second, given the hand grasp pose and object point clouds as inputs, we train another network that predicts the contact map on the object. We name this model the ContactNet. The key role of the ContactNet is to provide supervision to finetune GraspCVAE during test time when no ground truth is available. We design a self-supervised consistency task, which requires the hand contact points produced by the GraspCVAE to be consistent and overlapped with the object contact map predicted by the ContactNet. We use this self-supervised task to perform test-time adaptation which finetunes the GraspCVAE to generate a better human grasp. This adaptation approach can be applied on each single test instance. We emphasize that this procedure does not require any extra outside supervision and it can flexibly adapt to different inputs by resuming to the model before adaptation.

We evaluate our approach on multiple datasets include Obman~\cite{hasson2019learning}, HO-3D~\cite{Hampali2019HOnnotateAM} and FPHA~\cite{GarciaHernando2018FirstPersonHA} datasets. We show that by utilizing the novel objectives based on the contact consistency constraints in training time, we achieve significant improvements on human grasps generation against state-of-the-art approaches. More interestingly, by optimizing with the proposed self-supervised task during test time, it generalizes and adapts our model to unseen and out-of-domain objects, leading to the large performance gain.

Our contributions of this paper include: (i) Novel hand-object contact consistency constraints for learning human grasp generation; (ii) A new self-supervised task based on the consistency constraints which allows the generation model to be adjusted even during test time; (iii) Significant improvement on grasp generation for both in-domain and out-of-domain objects.

\section{Related Works}
\vspace{-0.05in}

\textbf{Hand-object interaction.} 
Modeling and analysing hand-object interaction is an active field of study with two main paradigms: joint estimating hand-object poses simultaneously during interaction~\cite{GarciaHernando2018FirstPersonHA, tekin2019h+, hasson2019learning, Doosti2020HOPENetAG, oberweger2019generalized, Rogez2015UnderstandingEH, Hussain2020FPHAAffordAD, liu2021semi} and studying from multi-modal hand-object interaction representations~\cite{Glauser2019InteractiveHP, Puhlmann2016ACR, brahmbhatt2019contactdb, brahmbhatt2020contactpose, taheri2020grab, Sundaram2019LearningTS, Ehsani2020UseTF}.  To perform hand-object pose estimation, Tekin \textit{et al.}~\cite{tekin2019h+} proposed a 3D detection framework, where the hand-object poses are predicted by two output grids without explicit interaction between them. On the contrary, Hasson~\textit{et al.}~\cite{hasson2019learning} leveraged the hand-centric physical constraints for modeling the interaction between hand-object to avoid penetration. Inspired by this work, we also applied explicit constraints for grasp generation. 

Another paradigm of study is to analyze the forces on hand and contact regions on objects from the multi-modal data. For example,  Sundaram \textit{et al.}~\cite{Sundaram2019LearningTS} introduced a scalable tactile glove, and utilized the touching information for object classification, while
Glauser \textit{et al.}
~\cite{Glauser2019InteractiveHP} leveraged it for a more difficult hand pose estimation task. Instead of using the tactile sensors on hand, Brahmbhatt \textit{et al.}~\cite{brahmbhatt2019contactdb, brahmbhatt2020contactpose} proposed to use thermal cameras to capture object contact maps, which reflects the object common contact regions after grasping. Inspired by this work, Taheri \textit{et al.}~\cite{taheri2020grab} further built a GRAB dataset, which not only captures the contact map from hand, but also takes the whole human body into consideration. This line of research motivates us to go beyond modeling hand-centric grasp generation, and explore the object-centric contact map by designing an object-centric loss to encourage the common contact regions on object to be touched by the hand.

\textbf{Grasp generation.} 
Generating human grasp is very challenging due to the higher degree-of-freedom of the human hand~\cite{karunratanakul2020grasping, corona2020ganhand, taheri2020grab, Brahmbhatt2019ContactGraspFM, Hussain2020FPHAAffordAD}.
To generate a realistic grasp, Karunratanakul \textit{et al.}~\cite{karunratanakul2020grasping} proposed an implicit representation for modeling the joint distribution of hand-object shape. Instead of implicit representation, our work is more related to work by Brahmbhatt \textit{et al.}~\cite{Brahmbhatt2019ContactGraspFM} which made use of the object contact maps to filter multiple generated grasps from GraspIt!~\cite{Miller2004GraspitAV}. However, the contact maps are taken as a constraint rather than a learning target in this grasp generation framework. 
In our work, we leverage the consistency between hand-object contact regions as training targets with the use of object contact map. Moreover, a self-supervised task is also designed for adjusting generated grasps using the contact maps at test-time.

\begin{figure*}[t]
\centering
\includegraphics[width=17.4cm]{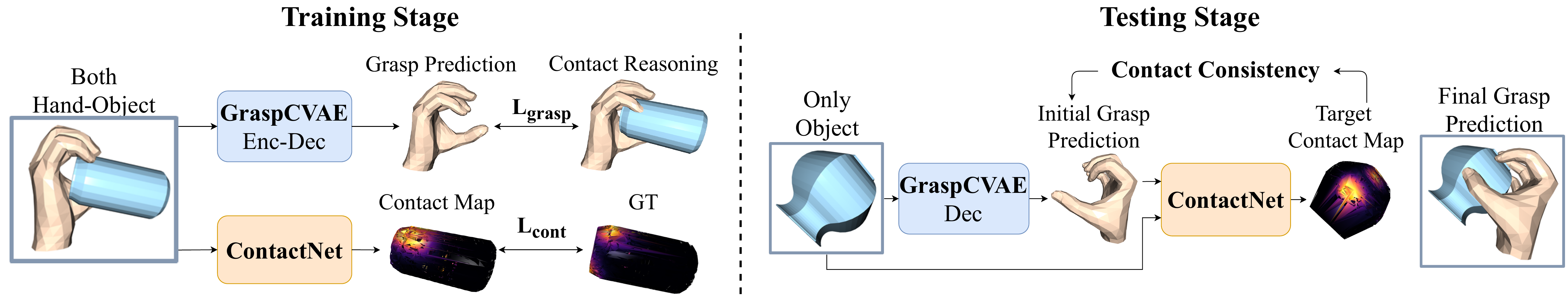}
\vspace{-0.25in}
\caption{\small{The \textit{different} usage of proposed networks in training and testing. \textbf{Left}: During training, the two networks learn generating human grasps and predicting object contact map separately on ground truth data. \textbf{Right}: At test-time, the two networks are unified in a cascade manner.
A initial grasp is predicted by the GraspCVAE \textit{decoder}, and it is inputted together with the object into ContactNet for predicting a target contact map.
We then leverage the contact consistency between outputs of the two networks for adjusting the initial grasp, where the target contact map serves as a self-supervision signal.}}
\vspace{-0.15in}
\label{fig:main}
\end{figure*}

\textbf{Affordance prediction.} Predicting scene and object affordance plays an important role in visual understanding~\cite{Li2019PuttingHI, Wang2019GeometricPA, Cao2020LongtermHM, Corona2020GanHandPH, Williams2019AnalysisOA, Do2018AffordanceNetAE, Kim2014SemanticLO, Fang2018Demo2VecRO, wang2020synthesizing}. For example, Corona \textit{et al.}~\cite{Corona2020GanHandPH} proposed a novel dataset and a generative network for learning grasps of multiple on-table objects. Williams \textit{et al.}~\cite{Williams2019AnalysisOA} leveraged object affordance prediction for real-world robotic manipulation. Inspired by these works, our goal is to generate grasps by learning the object affordance with ensuring the perceptual naturalness and physical plausibility at the same time. Different from the previous works, our method also allows better generalization of grasping out-of-domain objects with the help of the proposed self-supervised task.

\textbf{Learning on test instances.} Improving the generalization ability of neural networks is one of the most important problem in machine learning~\cite{Chen2011CoTrainingFD, Long2016UnsupervisedDA, Csurka2017DomainAF, Li2018DeepDG, Muandet2013DomainGV}. Recent research has started tackling the problem by leveraging self-supervision at test-time~\cite{Jain2011OnlineDA, Shocher2018ZeroShotSU, Bau2019SemanticPM, Sun2019TestTimeTW, Kalal2012TrackingLearningDetection, Mullapudi2019OnlineMD}. 
For example, Shocher \textit{et al.}~\cite{Shocher2018ZeroShotSU} proposed a self-supervised super-resolution framework where the network is only trained at test-time by up and downscaling \textbf{a single test example}. Sun \textit{et al.}~\cite{Sun2019TestTimeTW} extended the test-time adaptation idea to more general applications with a joint training framework of an image recognition task and a self-supervised task. At test-time, the network can be adjusted to a single test image by tuning the self-supervised objective. While this approach is intriguing, it is still unclear how the self-supervised objective can affect the main task objective. Inspired by this work, our approach also leverage self-supervision on a single instance for test-time adaptation. Different from~\cite{Sun2019TestTimeTW}, our self-supervised task is directly optimizing the main goal of generating better human grasps, which ensures the performance gain.

\section{Approach}

Our goal is generating hand meshes as human grasps given object point clouds as inputs. The generated hand mesh not only needs to be presented in a natural and realistic way, but it should also hold the object tightly in a physically plausible manner. We emphasize that ensuring reasonable contact between the object and synthesized hand is the key to get high-quality and stable human grasps.

To deal with this problem, we utilize both hands and object contact information and make sure they are consistent with each other, as summarized in Fig.~\ref{fig:main}. 
We propose two networks, a generative GraspCVAE to synthesize grasping hand mesh, and a deterministic ContactNet for modeling the contact regions on the object. 


\textbf{Training Stage.} As shown on the left side of Fig.~\ref{fig:main}, we optimize these two networks using ground-truth supervision \textit{separately} to learn grasp generation and predicting object contact maps. 
In this stage, the inputs of GraspCVAE are \textit{both} of hand and object, and GraspCVAE learns to synthesizing grasps in the \textit{hand reconstruction} paradigm, where \textit{both} of the its encoder and decoder will be used. Note that this follows the standard procedure in Conditional Variational Auto-Encoder (CVAE)~\cite{Sohn2015LearningSO}. To train the GraspCVAE, we propose two novel losses to ensure the hand-object contact consistency: one loss forcing the prior hand contact vertices to be close to the object surface, and another loss encouraging the object common contact regions to be touched by hand at the same time. The object and generated hand will find mutual agreement on the form of contact with the two losses during training.

\begin{figure*}[tb]
	\centering
        \begin{minipage}{0.75\textwidth}
        \includegraphics[width=\textwidth]{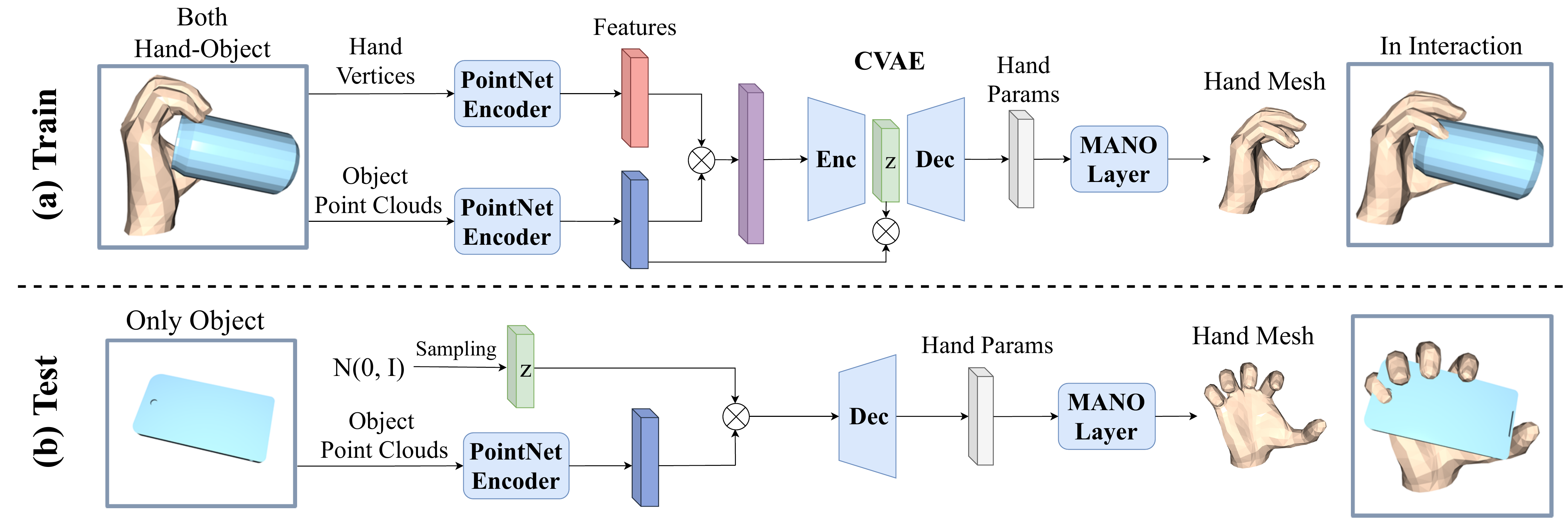}\vspace{-0.1in}
   \caption{\small{The architecture of GraspCVAE. (a) In training, it takes both hand-object as input to predict a hand mesh for grasping the object in a hand reconstruction manner using both of its \textit{encoder-decoder}; (b) At test-time, its \textit{decoder} generates grasps by conditioning only on object information as input. $\bigotimes$ means concatenation.}}
    \label{fig:GraspCVAE}
		\end{minipage}
        \hspace{0.1in}
        \begin{minipage}{0.2\textwidth}
        \includegraphics[width=\textwidth]{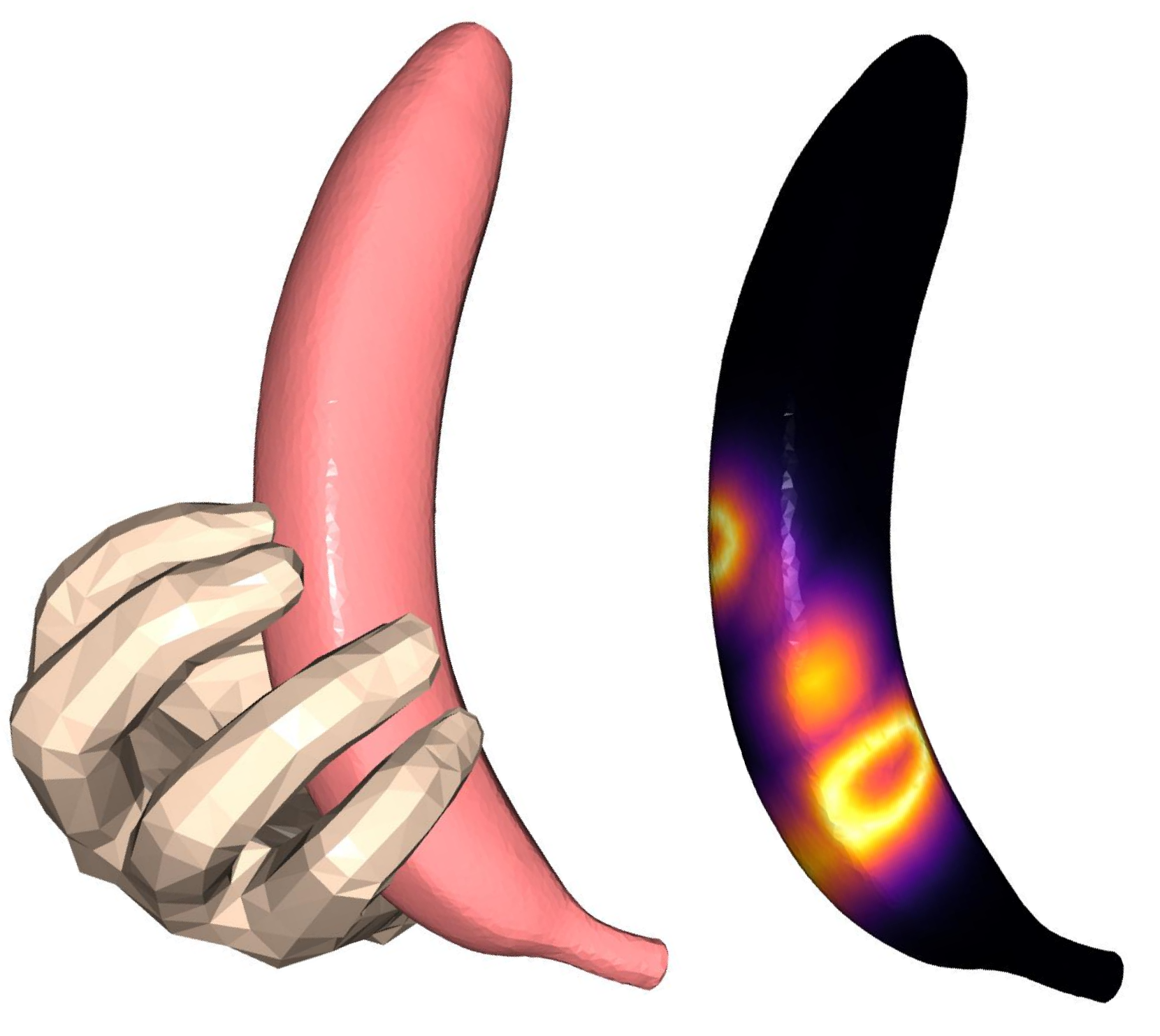}\hspace{-0.6mm}\vspace{-0.1in}
       \caption{\small{An example of contact map, brighter regions have larger scores. Because the MANO model does not have soft tissue, the underformable fingertips usually penetrate into object surface slightly.}}
    \label{fig: cmap}
		\end{minipage}
    \vspace{-0.2in}
\end{figure*}

\textbf{Testing Stage.} As shown on the right side of Fig.~\ref{fig:main}, we unify the two networks and design a self-supervised task by leveraging the consistency between their outputs. Given a test object, we first generate an initial grasp from the GraspCVAE decoder (without the encoder). Different from the training stage, the reconstruction target -- grasp is not provided in testing~\cite{Sohn2015LearningSO}. Then, the generated grasp is forwarded together with the object to the ContactNet to predict a target contact map. 
Since ContactNet is trained with ground truth data, where penetration between hand-object does not exist and hand fingers are touching the object surface closely, it will model the pattern of the ideal hand-object contact. During testing, the predicted contact map from ContactNet will tend to contain the ideal contact pattern.
We use the predicted contact map from the ContactNet as a target for finetuning and optimizing the grasps generated by GraspCVAE. If the grasp is predicted correctly from GraspCVAE, the object contact region from the predicted grasp should be consistent with the target object contact map.
We use this consistency as a self-supervision signal to adapt grasps generated by GraspCVAE during test-time.

In the following, we will first introduce the individual framework for GraspCVAE and ContactNet, and then the test-time contact reasoning with both networks for better adaptation to the new objects.

\subsection{Learning GraspCVAE}
\label{sec: graspcvae}
\vspace{-0.05in}

\paragraph{Usage and architecture} The GraspCVAE is a Conditional Variational Auto-Encoder (CVAE)~\cite{Sohn2015LearningSO} based generative network, which uses conditional information to control generation. For GraspCVAE, the conditional information is the object. We follow~\cite{kingma2013auto, Sohn2015LearningSO} to use the GraspCVAE: In training, \textit{both} of encoder and decoder of GraspCVAE is used to learn the grasp generation task in a hand reconstruction manner by taking in \textit{both} of hand-object as input; At test-time, \textit{only} its decoder is used to generate human grasp of a object with \textit{only} the 3D object as the input (without using the grasp for input). The network architecture is shown in Fig.~\ref{fig:GraspCVAE}.

\textbf{During training}, as shown in top row of Fig.~\ref{fig:GraspCVAE}, given two point clouds of the hand $\mathcal{P}^{h} \in \mathbb{R}^{778 \times 3}$ and the object $\mathcal{P}^{o} \in \mathbb{R}^{N \times 3}$ (where $N$ is the number of points) as inputs, we use two separate PointNets~\cite{Qi2017PointNetDL} to extract their features respectively, denoted as $\mathcal{F}^h, \mathcal{F}^o \in \mathbb{R}^{1024}$. These two features are then concatenated as $\mathcal{F}^{ho}$ for the GraspCVAE encoder inputs. The outputs of the encoder are the mean $\mu \in \mathbb{R}^{64}$ and variance $\sigma^2 \in \mathbb{R}^{64}$ of the posterior Gaussian distribution $Q(z | \mu, \sigma^2)$~\cite{kingma2013auto}. 
To reconstruct the hand, we first sample a latent code $z$ from the distribution and the posterior distribution ensures the sampled latent code $z$ is in correspondence with the input hand-object.

The decoder takes the concatenation of latent code $z$ and the object feature $\mathcal{F}^o$ as input to reconstruct a hand mesh,
which is represented by a differentiable MANO model~\cite{romero2017embodied}. The MANO model is parameterized by the shape parameter $\beta \in \mathbb{R}^{10}$ for person-specific hand shape, as well as the pose parameter $\theta \in \mathbb{R}^{51}$ for the joint axis-angles rotation and root joint translation. Given the predicted parameters $(\hat{\beta}, \hat{\theta})$ from the decoder, the MANO model forms a differentiable layer which outputs the shape of the hand with $\hat{\mathcal{M}} = (\hat{\mathcal{V}} \in \mathcal{R}^{778 \times 3}, \hat{F})$, where $\hat{\mathcal{V}}, \hat{F}$ denotes the mesh vertices and faces. Both the encoder and decoder in GraspCVAE are Multi-Layer Perceptrons (MLP).

\textbf{During testing}, as shown in the bottom row of Fig.~\ref{fig:GraspCVAE}, we only utilize the decoder from the GraspCVAE for inference. Given only the extracted object point cloud feature $\mathcal{F}^o$ and a latent code $z$ randomly sampled from a Gaussian distribution as inputs, the decoder will generate the parameters for the MANO model which leads to the hand mesh output. 

Given this architecture, we then introduce the training objectives as follows. We will first introduce the baseline objectives and then two novel losses which encourage the hand-object contact consistency.

\begin{figure*}[tb]
	\centering
        \begin{minipage}{0.25\textwidth}
        \includegraphics[width=\textwidth]{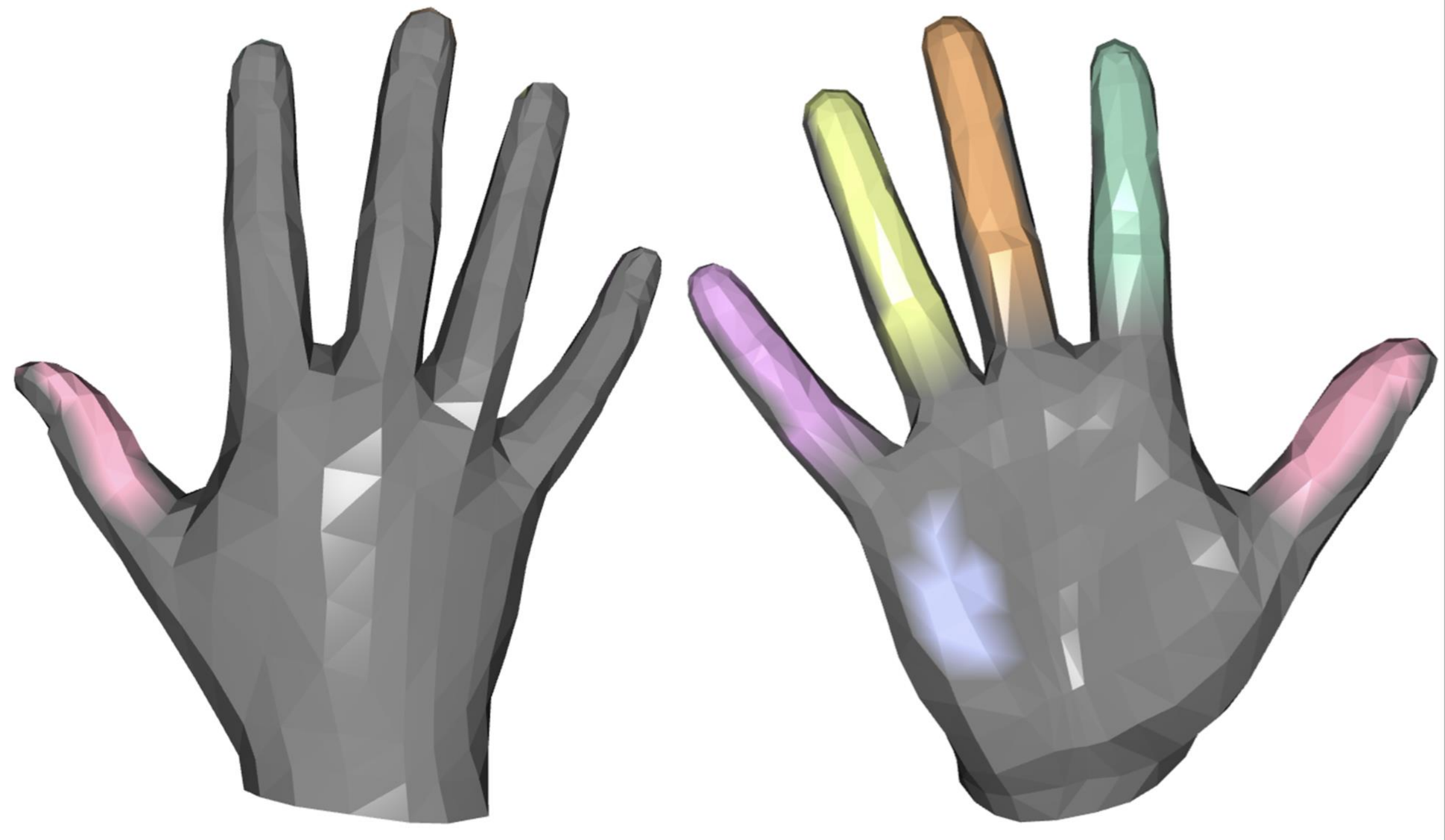}\hspace{-0.6mm}\vspace{-0.03in}
   \caption{\small{Six hand prior contact regions are shown in color.}}
    \label{fig:prior_function}
		\end{minipage}
		\hspace{0.1in}
        \begin{minipage}{0.7\textwidth}
        \includegraphics[width=\textwidth]{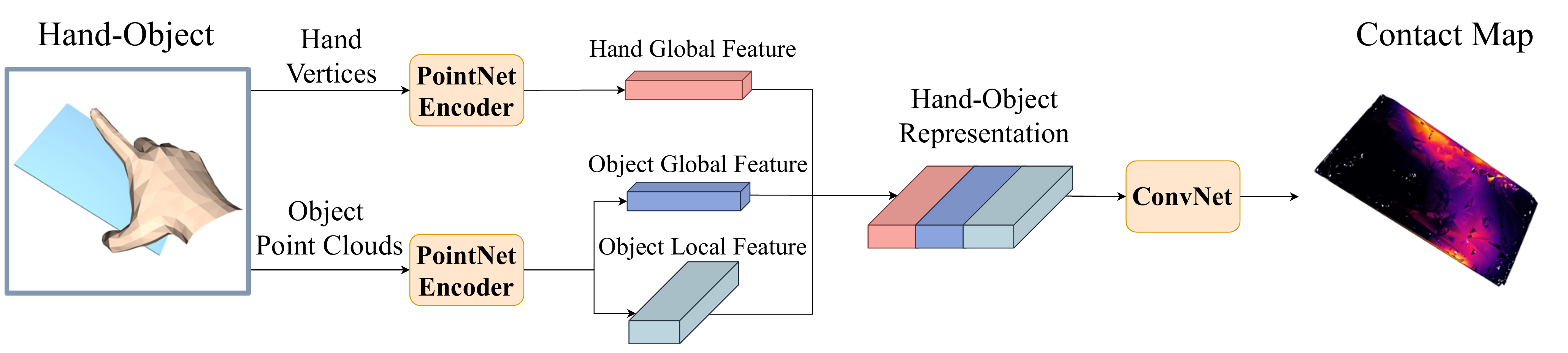}\vspace{-0.1in}
   \caption{\small{Architecture of ContactNet. It extracts local per-point feature of object point cloud, and concatenate it with global hand-object feature for predicting contact map.}}
\label{fig:contactnet}
		\end{minipage}
    \vspace{-0.15in}
\end{figure*}

\vspace{-0.15in}
\paragraph{Baseline}
The first objective for the baseline model is mesh reconstruction error, which is defined on both the vertices of the mesh as well as the parameters of the MANO model. We adopt the $L_2$ distance to compute the error. We denote the reconstruction loss between the predicted vertices and the ground-truths as $L_\mathcal{V} = \vert\vert \hat{\mathcal{V}} - \mathcal{P}^h \vert\vert^2_2$. The losses on MANO parameters are defined in a similar way with $L_\theta$ and $L_\beta$. The reconstruction error can be represented as$
    L_{\mathcal{R}} = \lambda_\mathcal{V} \cdot  L_\mathcal{V} + \lambda_\theta \cdot L_\theta + \lambda_\beta \cdot L_\beta\
$, where $\lambda_\mathcal{V}, \lambda_\theta$ and $\lambda_\beta$ are constants balancing the losses. 

Following the training of VAE~\cite{kingma2013auto}, we define a loss enforcing the latent code distribution $Q(z | \mu, \sigma^2)$ to be close to a standard Gaussian distribution, which is achieved by maximizing the KL-Divergence as
$
    L_{\mathcal{KLD}} = - KL(Q(z | \mu, \sigma^2) \vert\vert \mathcal{N}(0, I))\ .
$

We also encourage the grasp to be physically plausible, which means the object and hand should not penetrate into each other. We denote the object point subset that is inside the hand as $\mathcal{P}^o_{in}$, then the penetration loss is defined as minimizing their distances to their closest hand vertices $
    L_{penetr} = \frac{1}{\vert\mathcal{P}^o_{in}\vert} \sum_{p\in\mathcal{P}^o_{in}} \min_i \vert\vert p-\hat{V}_i\vert\vert_2^2\ .
$
In a short summary, the loss for training the baseline is:
\begin{align}
    L_{base} = L_{\mathcal{R}} + \lambda_{\mathcal{KLD}} \cdot L_{\mathcal{KLD}} + \lambda_{p} \cdot L_{penetr} \ ,
\end{align}
where $\lambda_\mathcal{KLD}$ and $\lambda_p$ are constants balancing the losses.

\vspace{-0.15in}
\paragraph{Reasoning Contact in Training}
There are two potential challenges in the baseline framework: First, the losses in the baseline model ignore physical contact between the hand-object, which cannot ensure the stability of the grasp; Second, grasp generation is multi-modal and the ground-truth hand pose is not the only answer. To tackle these challenges, we design two novel losses from both the hand and the object aspects to reason plausible hand-object contact and find the mutual agreement between them.

\textbf{Object-centric Loss.} From the object perspective, there are regions that are often contacted by human hand. We encourage the human hand to get close to these regions using the object-centric loss. Specifically, from the ground-truth hand-object interaction, we can derive the object contact map $\Omega \in \mathbb{R}^{N}$ by normalizing the distance $\textbf{D}(\mathcal{P}^o)$ between all object points and their nearest hand prior vertice with function $f(\cdot)$, where $f(\textbf{D}(\mathcal{P}^o)) = 1-2\cdot (\text{Sigmoid}(2\textbf{D}(\mathcal{P}^o))-0.5)$. An example is shown in Fig.~\ref{fig: cmap}. The distances are in center-meter and contact map scores are in $[0,1]$. This normalization helps the network focus on object regions close to the hand. Then we force the object contact map $\hat{\Omega}$ computed from the generated hand to be close to the ground truth $\Omega$, using loss
\begin{align}
\vspace{-0.1in}
    L_{\mathcal{O}} = \vert\vert \hat{\Omega} - \Omega \vert\vert^2_2,\ \Omega = f(\textbf{D}(\mathcal{P}^o)).
    \vspace{-0.2in}
\end{align}

\textbf{Hand-centric Loss.} We define the prior hand contact vertices $\mathcal{V}^p$ as shown in Fig.~\ref{fig:prior_function}, motivated by~\cite{hasson2019learning, brahmbhatt2019contactdb}. Given the predicted locations of the hand contact vertices, we then take the object points nearby as possible points to contact. Specifically, for each object point $\mathcal{P}^o_i$, we compute the distance $\textbf{D}(\mathcal{P}^o_i) = \min_j \vert\vert \mathcal{V}^p_j - \mathcal{P}^o_i \vert\vert^2_2$, and if it is smaller than a threshold, we take it as the possible contact point on the object. Our hand-centric objective is to push the hand contact vertices close to the object as,
\begin{align}
\vspace{-0.1in}
    L_{\mathcal{H}} = \begin{matrix} \sum_{i} \textbf{D}(\mathcal{P}^o_i) \end{matrix},\  for\ all\ \textbf{D}(\mathcal{P}^o_i) \leq \mathcal{T}
\end{align}
for all the possible contact points on the object, where $\mathcal{T}=1\ cm$ is the threshold.
The final loss combining the two novel losses above is,
\begin{align}
    L_{grasp} = L_{baseline} + \lambda_{\mathcal{H}} \cdot L_{\mathcal{H}} + \lambda_{\mathcal{O}} \cdot L_{\mathcal{O}} \ ,
    \vspace{-0.1in}
\end{align}
where $\lambda_{\mathcal{H}}$ and $\lambda_{\mathcal{O}}$ are constants balancing the losses. Intuitively, the $L_{\mathcal{O}}$ generally answers the question \textit{Where to grasp?}
and does not specify which hand part should be close to the object contact regions. And $L_{\mathcal{H}}$ is used to find the answer of \textit{Which finger should contact?} dynamically.
During training, with the two proposed losses, the hand contact points and object contact region will reach mutual agreement and be consistent to each other for generating stable grasps.

\subsection{Learning ContactNet}

We propose another network, the ContactNet, to
model the contact information between the hand-object as shown in Fig.~\ref{fig:contactnet}.
The inputs are hand and object point cloud, and the output is the object contact map denoted as $\Omega^c \in \mathbb{R}^{N}$ for $N$ object points. We use two PointNet encoders to extract hand and object feature maps. Since we need to predict the contact score for each point ($\Omega^c_i$ should be the score for $\mathcal{P}^o_i$), we utilize the per-point object local feature $\mathcal{F}^s \in \mathbb{R}^{N \times 64}$ of the PointNet encoder to ensure this correspondence. We also make use of hand and object global feature by first summing them then duplicate $N$ times and concatenate it with the object local feature for a to dimension $\mathbb{R}^{N \times 1024}$ feature maps.
Given these features, we apply four layers of 1-D convolution on top to regress the object contact map activated by the sigmoid function. The loss for training is the $L_2$ distance between the predicted contact map $\Omega^c$ and ground-truth $\Omega$ as, $L_{cont} = \vert\vert \Omega^c - \Omega \vert\vert^2_2$. 

During training time, the inputs for the ContactNet are directly obtained from the ground-truths.

\subsection{Contact Reasoning for Test-Time Adaptation}
\label{sec:TTA_method}
During testing, we unify the GraspCVAE and ContactNet in a cascade manner as shown on the right side of Fig.~\ref{fig:main}. Given the object point clouds as inputs, the GraspCVAE will first generate a hand mesh $\hat{\mathcal{M}}$ as the initial grasp. We compute its object contact map $\Omega_{\hat{\mathcal{M}}}$ correspondingly. Taking both the predicted hand mesh and object as inputs, the ContactNet will predict another contact map $\Omega^c$. If the grasp is predicted correctly, the two contact map $\Omega_{\hat{\mathcal{M}}}$ and $\Omega^c$ should be consistent. Based on this observation, we define a self-supervised consistency loss as $L_{refine} = \vert\vert \Omega_{\hat{\mathcal{M}}} - \Omega^c \vert\vert^2_2$ for fine-tuning the GraspCVAE. Besides this consistency loss, we also incorporate the hand-centric loss  $L_{\mathcal{H}}$ and penetration loss $L_{penetr}$ to ensure the grasp is physically plausible. We apply the joint optimization with all three losses on a \textit{single} test example as, 
\begin{align}
    L_{TTA} = L_{refine} + \lambda_{\mathcal{H}} \cdot L_{\mathcal{H}} + \lambda_{p} \cdot L_{penetr} .
\end{align}
We use this loss to update the GraspCVAE \textit{decoder}, and freeze other parts of the two networks.

\begin{figure*}[t]
\centering
\includegraphics[width=0.95\linewidth]{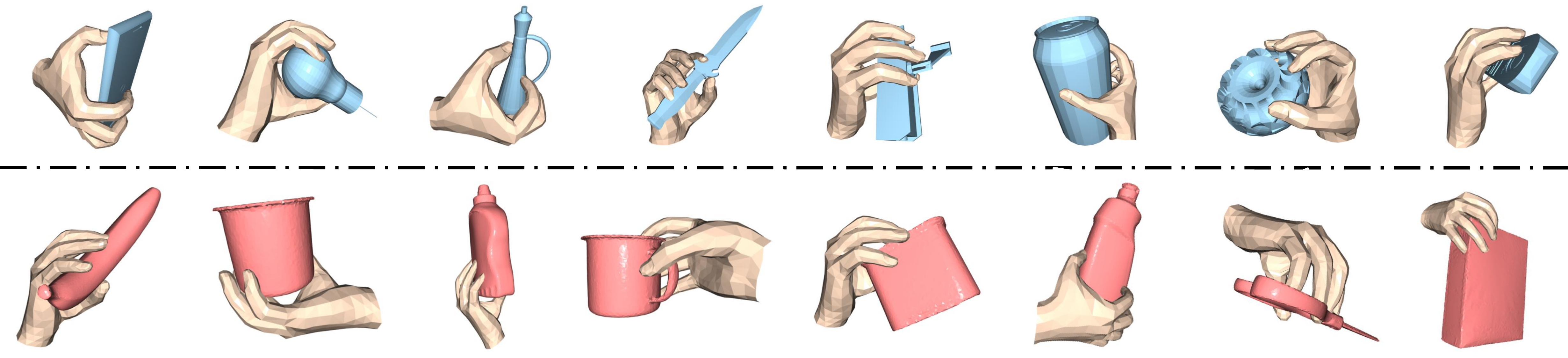}
\vspace{-0.15in}
\caption{\small{Visualization of generated grasps on \textcolor{obmanblue}{in-domain} objects from Obman dataset~\cite{hasson2019learning}, and  \textcolor{hored}{out-of-domain} objects from HO-3D dataset~\cite{Hampali2019HOnnotateAM}. See Fig.~\ref{fig: sup_obman} and Fig.~\ref{fig: sup_hoed_fpha} in Appendix for more results.}}
\label{fig:vis_grasp}
\vspace{-0.15in}
\end{figure*}

\begin{table*}[t]
\footnotesize
\tablestyle{5pt}{1.05}
\begin{tabular}{ll|ccc|ccc|ccc}
 & &\multicolumn{3}{c|}{Obman} & \multicolumn{3}{c|}{HO-3D} & \multicolumn{3}{c}{FPHA}
\\ 
 & & GT & GF~\cite{karunratanakul2020grasping} & Ours & GT & GF~\cite{karunratanakul2020grasping} & Ours & GT & GF~\cite{karunratanakul2020grasping} & Ours \\
\shline
Penetration & Depth ($cm$) $\downarrow$ & \textbf{0.01} & 0.56 & 0.46 & 2.94 & 1.46 & \textbf{1.05} & \textbf{1.17} & 2.37 & 1.58\\
& Volume ($cm^3$) $\downarrow$  & \textbf{1.70} &  6.05 & 5.12 & 6.08 & 14.90 & \textbf{4.58} & \textbf{5.02} & 21.9 & 6.37 \\\hline

Grasp Displace. & Mean ($cm$) $\downarrow$ & 1.66 & 2.07 & \textbf{1.52} & 4.31 & 3.45 & \textbf{3.21} & 5.54 & 4.62 & \textbf{2.55}\\
& Variance ($cm$) $\downarrow$ & $\pm$ 2.44 & $\pm$ 2.81 & $\pm$ \textbf{2.29} & $\pm$ 4.42 & $\pm$ 3.92 & $\pm$\textbf{3.79} & $\pm$ 4.38 & $\pm$ 4.48 & $\pm$ \textbf{2.22}\\\hline

Perceptual Score & $\{1, ..., 5 \}$ $\uparrow$ & 3.24 & 3.02 & \textbf{3.54} & 3.18 & 3.29 & \textbf{3.50} & 3.49 & 3.33 & \textbf{3.57}\\\hline

Contact & Ratio ($\%$) $\uparrow$ & \textbf{100} & 89.40 & 99.97 & 91.60 & 90.10 & \textbf{99.61} & 91.40 &  97.00 & \textbf{100}

\end{tabular}
\vspace{-0.1in}
\caption{\small{Results on Obman~\cite{hasson2019learning}, HO-3D~\cite{Hampali2019HOnnotateAM} and FPHA datasets~\cite{GarciaHernando2018FirstPersonHA} compared with ground truth (GT) and GF~\cite{karunratanakul2020grasping}. Best ones are in bold.}}
\label{tab: sota_compare}
\vspace{-0.15in}
\end{table*}

\section{Experiment}
We show qualitative results of generated grasps from our methods, and compare the qualitative performance with other methods in Sec.~\ref{sec:performance}. Then, we give ablation studies on the effectiveness of proposed novel losses during training and different Test-Time Adaptation (TTA) paradigms in Sec.~\ref{sec:ablation}.

\subsection{Implementation Details}
We sample $N=3000$ points on the object mesh as the input object point clouds.
In training, we use Adam optimizer and $LR=1e-4$ with 100 epochs, where the $LR$ is reduced half when model trained $30, 60, 80, 90$ epochs. Batch size is $128$. The loss weights are $\lambda_\beta=0.1$, $\lambda_\theta=0.1$, $\lambda_p=5$, $\lambda_{\mathcal{H}}=1500$ and $\lambda_{\mathcal{O}}=100$.
For Test-Time Adaptation, we use optimizer SGD with Momentum $0.8$, $LR=6.25\times10^{-6}$ is same as last epochs in training. For each sample, we use batch augmentations with batch size $32$. The loss weights are $\lambda_p=5$, $\lambda_{\mathcal{H}}=1$ and $\lambda_{\mathcal{O}}=5$.

\subsection{Datasets}
\textbf{Obman Dataset}~\cite{hasson2019learning} is a synthetic dataset, which includes hand-object mesh pairs. The hands are generated by a non-learning based method GraspIt!~\cite{Miller2004GraspitAV} and are parameterized by the MANO model. 2772 object meshes covering 8 classes of everyday
objects from ShapeNet~\cite{Chang2015ShapeNetAI} dataset are included. The model trained on this dataset will benefit from the diversified object models and grasp types. We train the two networks on this dataset as the initial model.

\textbf{HO-3D and FPHA Dataset}~\cite{GarciaHernando2018FirstPersonHA, Hampali2019HOnnotateAM} are two real datasets for studying hand-object interaction, and we use them for evaluating the generalization ability of our proposed framework. These two datasets collect video sequences annotated with object-hand poses. Because only a dozen of objects are included in these two datasets, they are not suitable for training the model. Besides, the objects in these two datasets have larger scales. We use the same split and data filtering of the two datasets with ~\cite{karunratanakul2020grasping}.

\begin{figure*}[t]
\centering
\includegraphics[width=\linewidth]{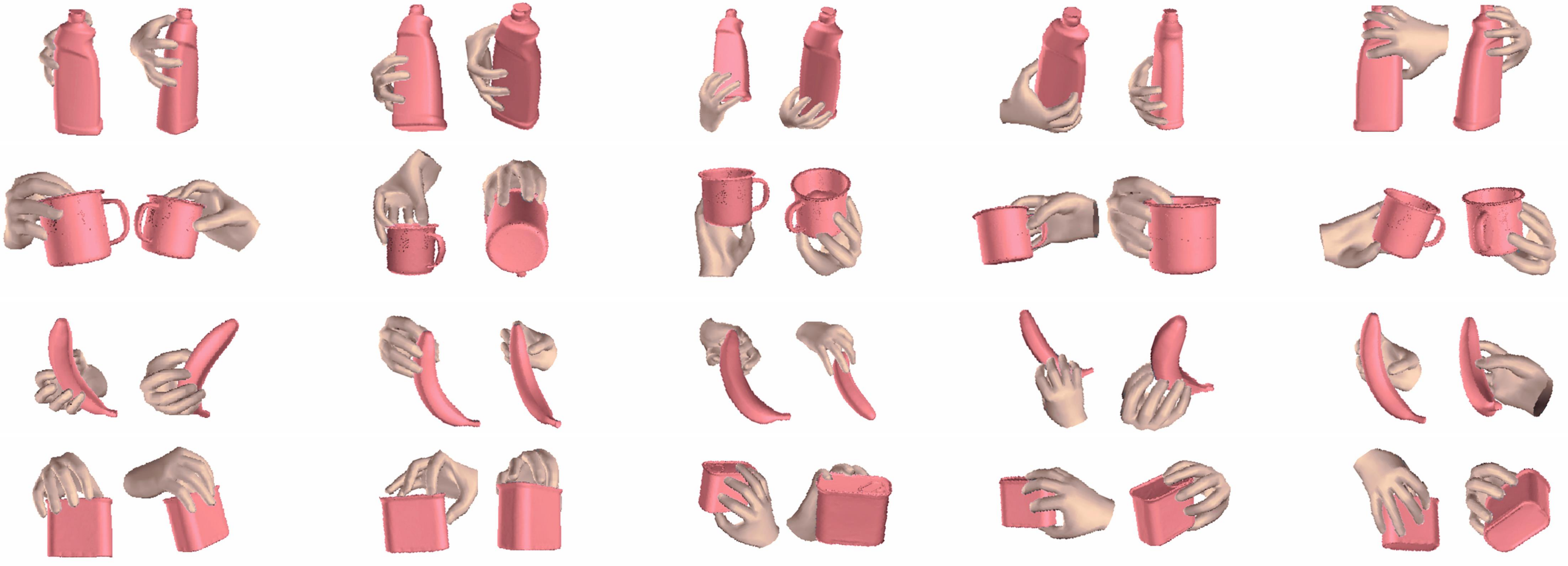}
\vspace{-0.25in}
\caption{\small{Diversity of generated grasps with examples on four \textcolor{hored}{out-of-domain} objects. We show 5 results for each object, where each example is shown in 2 views.}}
\label{fig: diversity}
\vspace{-0.1in}
\end{figure*}

\begin{table*}[t]
\tiny
\tablestyle{4pt}{1.05}
\begin{tabular}{l|cc|c|ccccc}
 &  \multicolumn{2}{c|}{Penetration $\downarrow$} & \multicolumn{1}{c|}{Grasp Displace. $\downarrow$} & \multicolumn{5}{c}{Contact $\uparrow$} 
\\ 
Loss & Depth & Volume  & Mean  $\pm$ Variance & Ratio ($\%$) & Obj Verts ($\%$) & Hand Verts ($\%$) & $\#$ Finger & CMap Score\\
\shline
$\mathcal{L}_{base}$ & \textbf{0.40} & \textbf{3.00} & 3.51 $\pm$ 3.70 & 97.69 & 3.58 & 8.20 & 2.97 & 6.42\\
\hline
\demph{+$\mathcal{L}_{H} (gt)$} & \demph{0.64} & \demph{4.90} & \demph{1.97 $\pm$ 2.83} & \demph{99.89} & \demph{6.07} & \demph{11.32} & \demph{3.67} & \demph{10.20}\\
+$\mathcal{L_H}$ & 0.48 & 4.85 & 1.72 $\pm$ 2.44 & 99.90 & 6.98 & 12.17 & \textbf{3.90} & 11.11\\
\hline
\demph{+$\mathcal{L_H}$+$\mathcal{L_O} (dist)$} & \demph{0.47} & \demph{4.72} & \demph{1.77 $\pm$ 2.65} & \demph{99.83} & \demph{6.80} & \demph{11.92} & \demph{3.82} & \demph{10.89}\\
+$\mathcal{L_H}$+$\mathcal{L_O}$ & 0.48 & 4.92 & \textbf{1.63 $\pm$ 2.43} & \textbf{99.94} & \textbf{7.16} & \textbf{12.17} & 3.87 & \textbf{11.24}\\
\end{tabular}
\vspace{-0.1in}
\caption{\small{Ablation study for proposed losses of GraspCVAE on the Obman~\cite{hasson2019learning}. To verify the effectiveness of each loss, we also compared each of them with a modified version shown in \demph{gray} additionaly.}}
\label{tab: ablation_graspcvae}
\vspace{-0.15in}
\end{table*}

\subsection{Evaluation Metrics}
\vspace{-0.05in}
\textbf{Penetration} is measured by penetration \textbf{depth} and \textbf{volume} between objects and generated grasps following~\cite{hasson2019learning}. We voxelize the hand-object meshes with voxel size $0.5\ cm$, and calculate the intersection shared by the two 3D voxels. 

\textbf{Grasp displacement} is used to measure the stability of the grasp.
To test stability, we put the object and generated grasp in a simulator following~\cite{Tzionas2016CapturingHI, hasson2019learning}. In general, the simulator calculates the motion of the object under the grasp.
Specifically, the simulator calculates forces on fingertips, which are has a positive correlation with the penetration volume on fingertips.
Then, it applies the calculated forces to hold the object against its gravity.
The grasp stability is measured by the displacement of the object's center of mass during a period in the simulation. In this period, the pose and location of hand is fixed. We measure the mean and variance of the simulation displacement for all test samples.
Examples with smaller simulation displacement have better grasp stability.
The correlation between the grasp stability and penetration of the grasp is discussed in Sec.~\ref{sec: penetration_stability}.

\textbf{Perceptual score} is utilized for evaluating the naturalness of generated grasps. We perform the perceptual study following ~\cite{karunratanakul2020grasping} with Amazon Mechanical Turk. 

\textbf{Hand-object contact metrics} are used for analyzing contact between hand-object. We calculate the sample-level hand-object contact ratio, individual object and hand contact points ratio, and the number of hand fingers contacting the object. We classify the contact status of a point by judging whether its distance to its nearest neighbor in the other point cloud is smaller than $0.5\ cm$. We also calculate the object contact map score, as $s = 100 \cdot \frac{\sum \Omega}{N} \in [0,100]$, which reflects the coverage area of grasps. Generally, larger contact areas can imply a better grasp, but this is not \textit{strictly} correct.

\subsection{Grasp Generation Performance}
\label{sec:performance}

\textbf{Qualitative results.} We first visualize generated grasps for different objects. Fig.~\ref{fig:vis_grasp} shows that our framework is able to generate stable grasps with natural hand poses on both in-domain and out-of-domain objects. By sampling different object poses for the same object as inputs, our model can generate diverse grasps. Fig.~\ref{fig: diversity} shows 5 different grasps generated by our model for each object in each row.  


\textbf{Quantitative results.}
The evaluation results on the three datasets are shown in Table~\ref{tab: sota_compare}.
We train the models on the \textbf{Obman} training set, and test on Obman testset. We also test the model trained from the Obman training set extensively on \textbf{HO-3D} and \textbf{FPHA} to demonstrate the generalization ability of our method. All results are evaluated after Test-Time Adaptation (TTA).
The objects in the Obman test set may overlap with its training set, while objects (with different poses) from HO-3D and FPHA are never seen in training.

On all of the three datasets, our framework shows significant improvement over the state-of-the-art approach~\cite{karunratanakul2020grasping} in \textit{both} of physical plausibility, grasp stability and perceptual score. 
And results on HO-3D and FPHA dataset imply that our model has a much stronger cross-domain generalization ability.
For instance, \cite{karunratanakul2020grasping} achieves reasonably good stability on HO-3D and FPHA but suffers from huge penetration (they are correlated). 
However, our model performs much better on both of the two metrics with a great balance.
Moreover, the perceptual scores of our framework on the three datasets are similar: $3.54$ for in domain objects in Obman, $3.50$ and $3.57$ for out-of-domain data in HO-3D and FPHA. This shows the quality of generated grasps on out-of-domain objects are close to the in-domain objects. 
Besides, our results are close to or even outperform the ground truth, especially for the stability and perceptual score. 
These results imply our method's capability to generating natural, physically plausible and steady grasps. 

\vspace{-0.01in}
\subsection{Ablation Study}
\label{sec:ablation}
\vspace{-0.03in}

We first perform ablation studies on Obman dataset~\cite{hasson2019learning} for evaluating the two proposed losses  $\mathcal{L_H}$ and $\mathcal{L_O}$. 
We then analyse different designs of ContactNet.
Finally, we compare different 
Test-Time Adaptation (TTA) paradigms on out-of-distribution HO-3D and FPHA~\cite{GarciaHernando2018FirstPersonHA, Hampali2019HOnnotateAM} dataset.

\begin{table}[t]
\tiny
\vspace{-0.1in}
\tablestyle{5pt}{1.05}
\begin{tabular}{l|ccc}
 & \multicolumn{3}{c}{Model}\\
 & object-only & h-o global & h-o global-local\\ \shline
 Error & 0.161 & 0.148 & \textbf{0.090} \\
\end{tabular}
\vspace{-0.1in}
\caption{\small{Ablation study of different ContactNet designs. The error is average contact map score absolute error of all object points between predictions and ground truth.}}
\label{tab: contactnet_ablation}
\vspace{-0.15in}
\end{table}

\begin{table*}[t]
\tiny
\tablestyle{4pt}{1.05}
\begin{tabular}{ll|cc|c|ccccc}
&  & \multicolumn{2}{c|}{Penetration $\downarrow$} & \multicolumn{1}{c|}{Grasp Displace. $\downarrow$} & \multicolumn{5}{c}{Contact $\uparrow$} 
\\ 
 & & Depth & Volume  & Mean  $\pm$ Variance & Ratio ($\%$) & Obj Verts ($\%$) & Hand Verts ($\%$) & $\#$ Finger & CMap Score\\
\shline
HO-3D~\cite{Hampali2019HOnnotateAM} & w/o TTA & \textbf{0.94} & \textbf{4.21} & 4.98 $\pm$ 4.48 & 86.63 & 3.41 & 8.78 & 3.11 & 5.65 \\
& TTA & 1.09 & 4.88 & 3.80 $\pm$ 4.20 & 92.31 & 4.37 & 10.83 & 3.58 & 7.13\\
& TTA-optm & 1.07 & 4.59 & 4.14 $\pm$ 4.31 & 91.45 & 4.32 & 10.97 & 3.68 & 6.78 \\
& TTA-noise & 1.12 & 4.98 & 4.22 $\pm$ 4.34 & 91.17 & 4.14 & 10.40 & 3.32 & 6.81\\
& TTA-online & 1.05 & 4.58 & \textbf{3.21 $\pm$ 3.79} & \textbf{99.61} & \textbf{4.66} & \textbf{11.55} & \textbf{3.88} & \textbf{7.80}\\
\hline
FPHA~\cite{GarciaHernando2018FirstPersonHA} & w/o TTA & \textbf{6.19} & \textbf{1.56} & 2.93 $\pm$ 2.70  &  100 & 4.71 & 13.78 & 4.47 & 7.67  \\
& TTA & 6.37 & 1.58 & \textbf{2.55 $\pm$ 2.22} & 100  & 4.64 & 13.95 & 4.56 & 7.67\\
& TTA-online & 6.31 & 1.69 & 2.77 $\pm$ 2.47 & \textbf{100} & \textbf{4.83} & \textbf{14.44} & \textbf{4.73} & \textbf{7.83}\\ 
\end{tabular}
\vspace{-0.1in}
\caption{\small{Results of different Test-Time Adaptation (TTA)  methods on out-of-domain HO-3D and FPHA dataset~\cite{Hampali2019HOnnotateAM, GarciaHernando2018FirstPersonHA}.}}
\label{tab: generalization}
\vspace{-0.1in}
\end{table*}

\vspace{-0.03in}
\subsubsection{GraspCVAE Training Objectives}
\vspace{-0.02in}
\label{sec:training_targets}
The results are shown in Table~\ref{tab: ablation_graspcvae}. 
With the hand-centric loss $\mathcal{L_H}$, the simulation displacement decreases and contact metrics grows significantly while the penetration grows slightly. 
After adding the object-centric loss $\mathcal{L_O}$, only object-related metrics, e.g. contact object vertices ratio and contact map score, and stability grows (displacement decreases). 
This implies that with $\mathcal{L_H}$, the $\mathcal{L_O}$ acts as a regularizer on the object contact regions to improves the grasp stability, which matches the design of this loss function.

We also verify the effectiveness of two losses by comparing each with a modified version.
First, we can force the fingers to touch the ground truth contact regions rather than finding them dynamically with $\mathcal{L_H}$. 
This loss is denoted as $\mathcal{L_H}(gt)$.
Experiments demonstrate that $\mathcal{L_H}$ is better in all metrics than $\mathcal{L_H}(gt)$. This implies that fitting the ground truth in the multi-solution grasp generation task may not be optimal.
Second, in the loss $\mathcal{L_O}$, we verify the effectiveness of using the contact map as the representation of hand-object distance. 
We experiment with directly minimize the residual between predicted and ground truth object-hand distances $\hat{\textbf{D}}$ and $\textbf{D}$ without normalizing them into contact maps. We call this loss as $\mathcal{L_O} (dist)$.
Experiments show that with $\mathcal{L_O} (dist)$, the performance even degenerates. The reason is that the $\mathcal{L_O}(dist)$ is contributed almost by hand-object point pairs with large distance, while $\mathcal{L_O}$ focus more on hand vertices close to object surface with the help of normalization.

\begin{figure*}[tb]
	\centering
        \begin{minipage}{0.45\textwidth}
        \vspace{0.15in}
        \includegraphics[width=\textwidth]{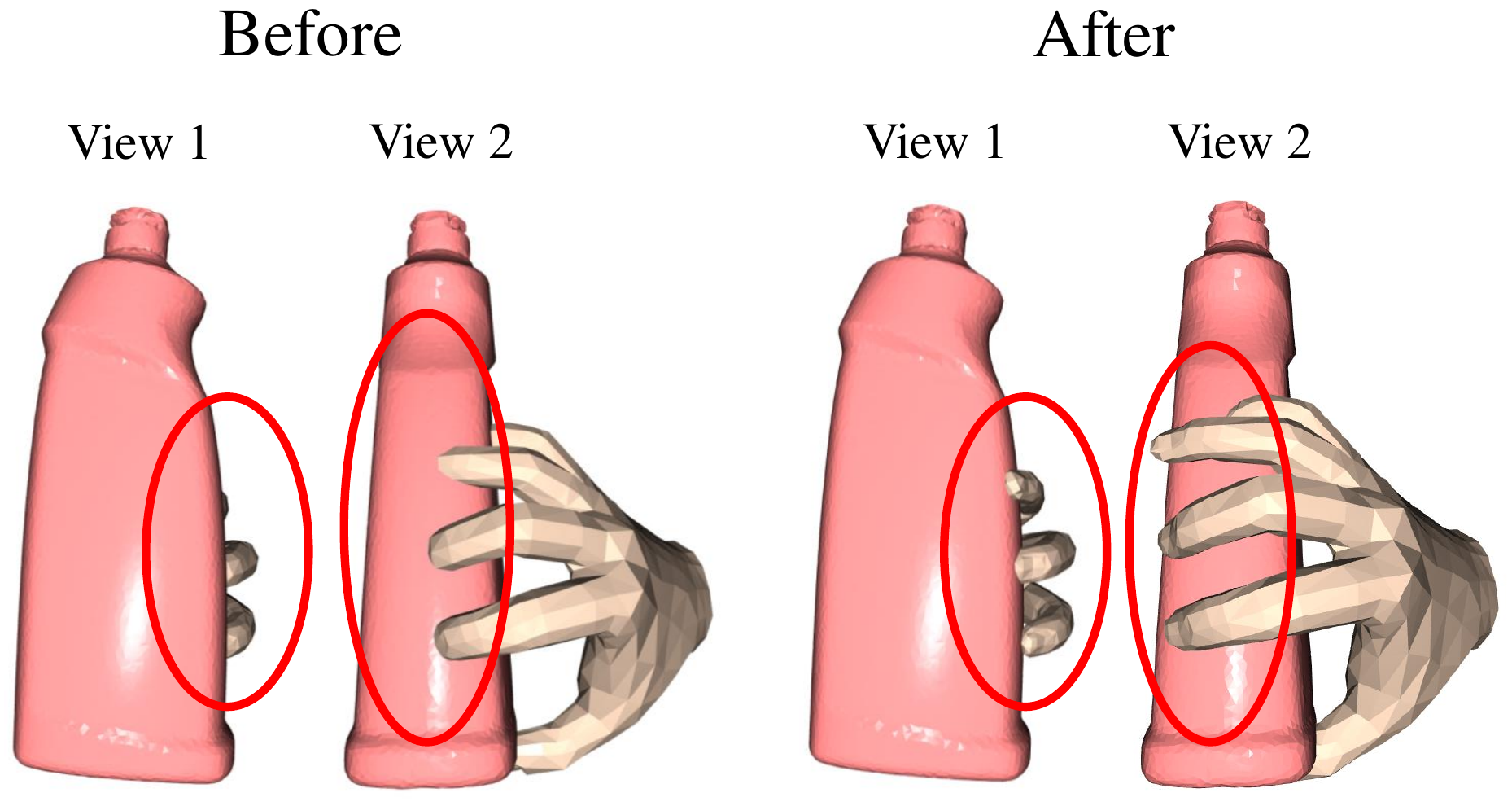}\vspace{0.05in}
   \caption{\small{Visualization of grasp before and after TTA. The penetration decreases on fingertips.}}
    \label{fig: TTA_hand}
		\end{minipage}
        \hspace{0.2in}
        \begin{minipage}{0.45\textwidth}
        \includegraphics[width=\textwidth]{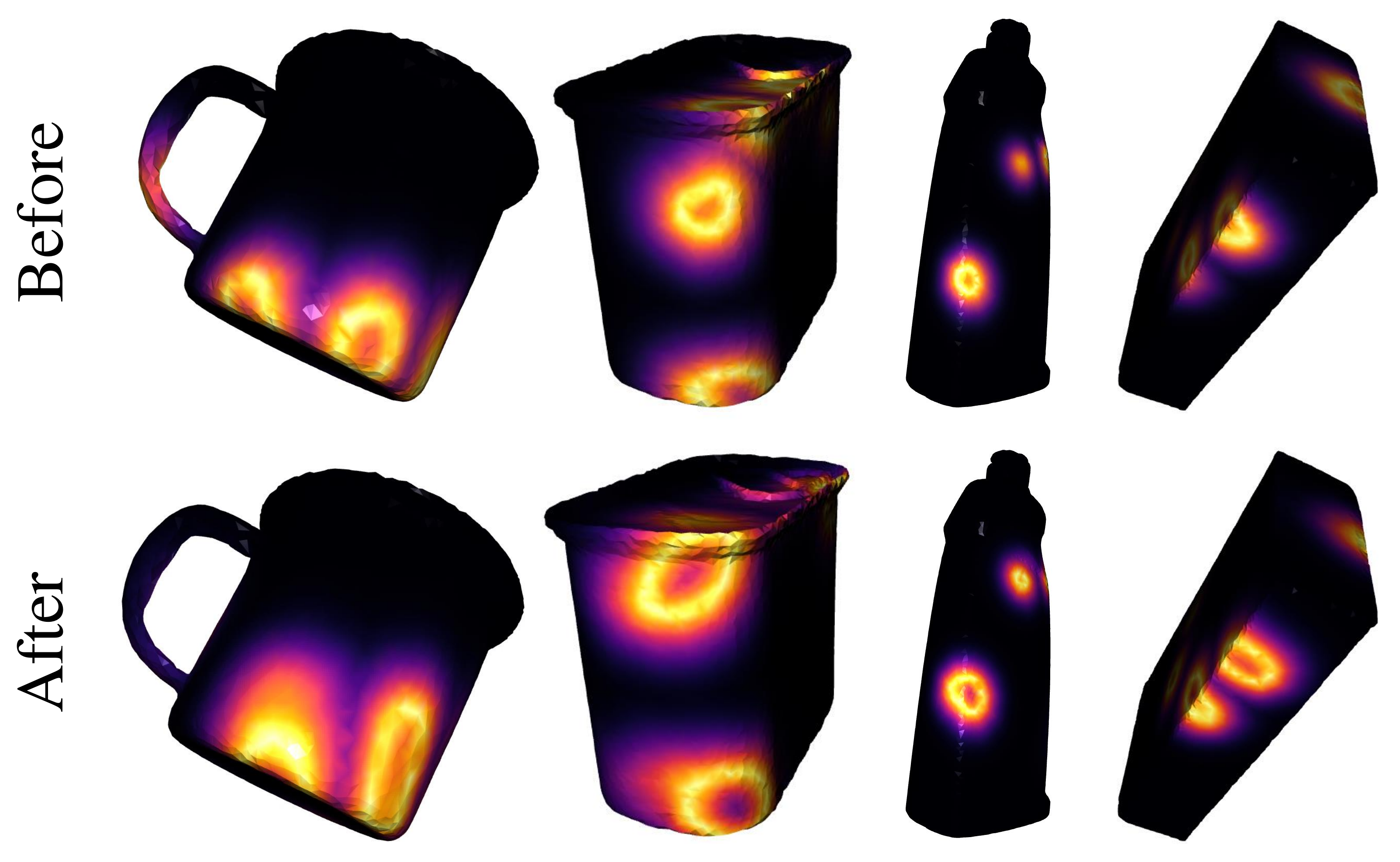}\hspace{-0.6mm}\vspace{-0.1in}
   \caption{\small{After TTA, object contact regions become larger, where the yellow circles on contact maps reflect the intersection rings between hand-object.}}
    \label{fig:TTA_object}
		\end{minipage}
\end{figure*}

\vspace{-0.1in}
\subsubsection{ContactNet Designs}
\label{sec:contactnet_designs}

We compare three different kinds of ContactNet designs, as shown in Table~\ref{tab: contactnet_ablation}. The first model (object-only) takes solely the object as input, while the second (h-o global) and third (h-o global-local) model take in both of hand and object. The difference between the latter two is to experiment whether using object local features helps predict the contact map by maintaining point permutation information.

Without the hand as an input, predicting object contact map is a very difficult one-to-many mapping. Considering only a small part of object points are in contact, the $0.161$ absolute error is actually huge.
Experiments also show that without object local features, the gain from adding the hand as one of the input is trivial. With object local features, the error reduces $0.07$ as a significant improvement of $50\%$.

\subsubsection{Test-Time Adaptation (TTA) for Generalization}
\label{sec:TTA}

We compare four different TTA paradigms:
\begin{itemize}
\vspace{-0.05in}
    \item TTA (offline): Learning-based TTA same as illustrated in Sec.~\ref{sec:TTA_method}, and network parameters are re-initialized before adapting each sample;
    \vspace{-0.1in}
    \item TTA-optm (offline): Optimization-based TTA, where the MANO parameters are directly optimized;
    \vspace{-0.1in}
    \item TTA-noise (offline): Learning-based TTA. When training ContactNet, the hand parameters are injected with random noise.
    The method is used to compare different methods to obtain the target contact map;
    \vspace{-0.1in}
    \item TTA-online: Learning-based TTA, and the network parameters are re-initialized only \textit{once} for each sequence.
    \vspace{-0.2in}
\end{itemize}

As shown in Table~\ref{tab: generalization}, on the two datasets, all TTA methods can improve the results.
There are three comparisons between the different methods. 
First, on the HO-3D dataset, the TTA and TTA-optm achieve comparable results because they are both offline methods using the same objective function. The results of the learning-based TTA are slightly better, which can be explained by that the network parameters serve as a regularization and make the adaptation more steady. 
Second, training with injected noise, we expect the ContactNet can learn to predict ideal contact maps as target in TTA by "correting" the noise. However, the results deteriorate compared with the one trained on perfect ground truth data. This can be explained by: (i) Injecting noise hurts learning contact maps; (ii) It is hard to match the random noise with the noise pattern in initially predicted grasps.
Third, the online version of TTA is much stronger than offline versions. Inspired by~\cite{Sun2019TestTimeTW, Jain2011OnlineDA, Mullapudi2019OnlineMD}, for the online TTA, the target of the TTA can be optimized continually with the help of network parameters, and the model can fit the test distribution better. With online updating, the stability grows and penetration depth decreases simultaneously, indicating that the network leans better hand-object contact. A huge improvement in contact ratio also verifies the point.
The improvement of online TTA on the FPHA dataset is not as big as on HO-3D dataset because the average video sequence length is $\frac{1}{20}$ of HO-3D dataset, so the learning target cannot be optimized continually.

To show the effectiveness of TTA for improving both the naturalness and stability of generated grasps, we further visualize the grasps and object contact maps before and after TTA. As shown Fig.~\ref{fig: TTA_hand}, after TTA, the hand penetration decreases with fingers closely contacting object surface. In Fig.~\ref{fig:TTA_object}, the object contact regions become larger, which indicates the grasps are more stable. 

\begin{figure}[t]
\centering
\includegraphics[width=0.9\linewidth]{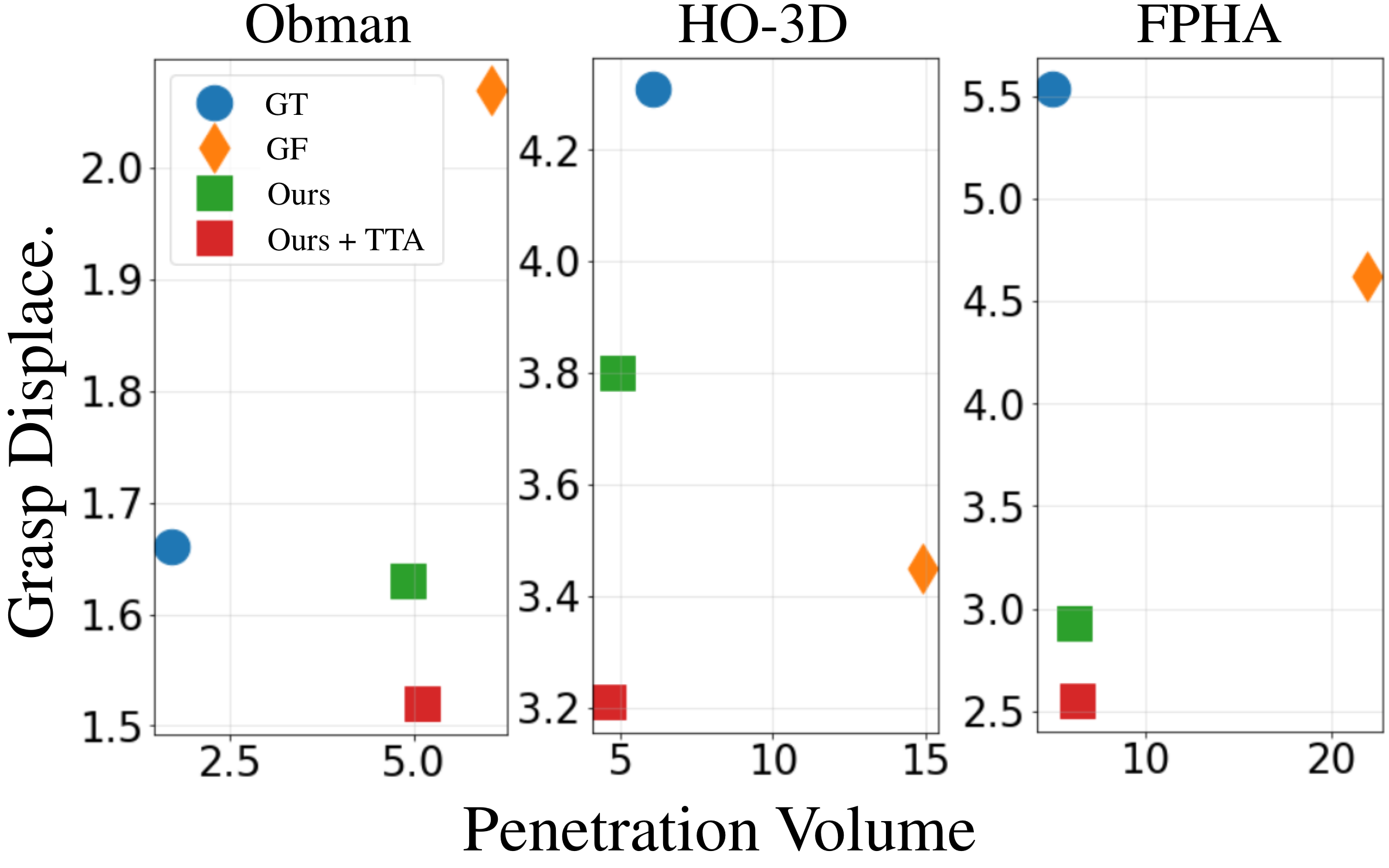}
\vspace{-0.05in}
\caption{\small{
Balance between penetration volume (x-axis) and grasp stability (y-axis, measured by simulation displacement) on the three datasets, compared with ground truth (GT) and Grasping Field (GF)~\cite{karunratanakul2020grasping}. Results close to the origin point are better, indicating a grasp has better stability (smaller simulation displacement) with smaller penetration.
}}
\label{fig:pene-sim}
\vspace{-0.15in}
\end{figure}

\subsection{Penetration Volume vs. Grasp Displace.}
\label{sec: penetration_stability}
Larger penetration volume can cause better grasp stability (reflected by smaller simulation displacement) during the simulation. 
However, ideal grasps should be with small penetration and simulation displacement simultaneously, rather than achieving reasonable stability by suffering from huge penetration volume.
Thus, we draw Fig.~\ref{fig:pene-sim} for demonstrating the balance between them on the three datasets.
Overall, our results are very close the origin point, which demonstrated our generated grasps has both small penetration and superior stability at the same time. With TTA, the results move vertically in the figure, indicating the TTA is able to increase grasp stability without magnifying the penetration at the same time.
Besides, the results are comparable to or even outperform the ground truth.

\section{Conclusion}
In this work, we propose a framework for generating human grasps given an object. To get natural and stable grasps, We reason the consistency of contact information between object and generated hand from two aspects: First, we design two novel training targets from the view of hand and object respectively, which helps them to find a mutual agreement on the form of contact. Second, we design two networks for grasp generation and predicting contact map respectively. We leverage the consistency between outputs of the two networks for designing a self-supervised task, which can be used at test-time for adapting generated grasps on novel objects. With the proposed method, we not only observe more natural and stable generated grasps, but also a strong generalization capability on cross-domain test inputs.

\vspace{1em}
{\footnotesize \textbf{Acknowledgements.}~This work was supported, in part, by grants from DARPA LwLL, NSF 1730158 CI-New: Cognitive Hardware and Software Ecosystem Community Infrastructure (CHASE-CI), NSF ACI-1541349 CC*DNI Pacific Research Platform, and gifts from Qualcomm and TuSimple.}

{\small
\bibliographystyle{ieee_fullname}
\bibliography{egbib}
}

\clearpage
\setcounter{section}{0}

\begin{center}
\textbf{\Large Appendix}
\end{center}

We provide more detailed information in the Appendix, including:
\vspace{-0.1in}
\begin{itemize}
    \item Network architectures;
    \vspace{-0.05in}
    \item Experiment and evaluation details;
    \vspace{-0.05in}
    \item More results with visualization.
\end{itemize}
\vspace{-0.15in}
\section*{Appendix A: Network Architectures}
We show structures of GraspCVAE and ContactNet in the following sections.

\subsection*{A.1. GraspCVAE}
\vspace{-0.15in}

\begin{table}[h]
\renewcommand{\arraystretch}{1.0}
\scriptsize
\tablestyle{7pt}{1.2}
\begin{tabular}{ccc}\shline
 Stage & Configuration & Output\\\shline
 \multirow{2}*{0} & Input Hand point cloud $\mathcal{P}^h$ & $778 \times 3$\\
  & Input Object point cloud $\mathcal{P}^o$ & $3000 \times 3$\\\shline
  \multicolumn{3}{c}{\textbf{3D Feature extraction}} \\ \hline
 \multirow{2}*{1} & \multirow{2}*{Extract feature with two PointNet encoders} & $1024$ ($\mathcal{F}^h$) \\ 
 & & $1024$ ($\mathcal{F}^o$) \\ \shline
 \multicolumn{3}{c}{\textbf{Calculating posterior distribution} (Input \textit{concat}($\mathcal{F}^h$, $\mathcal{F}^o$) )} \\ \hline
 \multirow{2}*{2} & \multirow{2}*{\shortstack{ CVAE encoder \\
 (fc-layers, $2048, 1024, 512, 256, 64$)}} & $64$ ($\mu$)\\ 
 & & $64$ ($\sigma^2$) \\\shline
 \multicolumn{3}{c}{\textbf{Latent code sampling}} \\ \hline
 3 & Sampling from calculated Gaussian & $64$ ($z$) \\ \shline 
 \multicolumn{3}{c}{\textbf{Hand mesh reconstruction} (Input \textit{concat}($\mathcal{F}^o$, $z$) )} \\ \hline
 \multirow{2}*{4} & \multirow{2}*{\shortstack{ CVAE decoder \\ (fc-layers, $1088, 1024, 256, 61$)}} & \multirow{2}*{\shortstack{$61$ \\ (param)}} \\ \\
 4 & MANO Layer & Hand mesh $\hat{\mathcal{M}}$ \\ \shline
\end{tabular}
\vspace{-0.05in}
\caption{Training time GrasCVAE architecture.}
\label{tab: train_graspcvae}
\end{table}

\vspace{-0.15in}
\begin{table}[h]
\scriptsize
\tablestyle{7pt}{1.2}
\begin{tabular}{ccc}\shline
 Stage & Configuration & Output\\\shline
 0 & Input Object point cloud $\mathcal{P}^o$ & $3000 \times 3$\\\shline
  \multicolumn{3}{c}{\textbf{3D Feature extraction}} \\ \hline
 1 & Extract feature with a PointNet encoder & $1024$ $(\mathcal{F}^o)$ \\ \shline
 \multicolumn{3}{c}{\textbf{Latent code sampling}} \\ \hline
 \multirow{2}*{2} & \multirow{2}*{\shortstack{ Random sampling in \\
 standard Gaussian}} & \multirow{2}*{$64$ $(z)$} \\ \\ \shline 
 \multicolumn{3}{c}{\textbf{Grasp Prediction} (Input \textit{concat}($\mathcal{F}^o$, $z$) )} \\ \hline
 \multirow{2}*{3} & \multirow{2}*{\shortstack{ CVAE decoder \\ (fc-layers, $1088, 1024, 256, 61$)}} & \multirow{2}*{\shortstack{$61$ \\ (param)}} \\ \\
 3 & MANO Layer & Hand mesh $\hat{\mathcal{M}}$ \\ \shline
\end{tabular}
\vspace{-0.05in}
\caption{Test-time GrasCVAE architecture.}
\vspace{-0.05in}
\label{tab: test_graspcvae}
\end{table}

Table~\ref{tab: train_graspcvae} and Table~\ref{tab: test_graspcvae} show the architecture of GraspCVAE during training and testing respectively. The input of the two phase are different. During training, the input is both of hand and object point cloud and we train the network in a hand reconstruction manner. During testing, the only input is the object point cloud, and the network generates human hand mesh for grasping the object. 

For training, we use two PointNet encoders to get features of hand and object point cloud as $\mathcal{F}^h$ and $\mathcal{F}^o$. Then, they are concatenated and sent to the CVAE encoder for predicting the posterior distribution $Q(z | \mu, \sigma^2)$. Then, a latent code $z$ is sampled from this distribution, and concatenated with the object feature $\mathcal{F}^o$ as the input of CVAE decoder for regressing the MANO parameters. In the end, the parameters pass the MANO layer, where the output is the generated hand mesh $\hat{\mathcal{M}}$.

For testing, the latent code $z$ is randomly sampled from the standard Gaussian distribution. Thus, we do not need the CVAE encoder and the hand point cloud.

\subsection*{A.2. ContactNet}
\vspace{-0.15in}
\begin{table}[h]
\scriptsize
\tablestyle{7pt}{1.2}
\begin{tabular}{ccc}\shline
 Stage & Configuration & Output\\\shline
 \multirow{2}*{0} & Input Hand point cloud $\mathcal{P}^h$ & $778 \times 3$\\
  & Input Object point cloud $\mathcal{P}^o$ & $3000 \times 3$\\\shline
  \multicolumn{3}{c}{\textbf{3D Feature extraction}} \\ \hline
 \multirow{3}*{1} & \multirow{3}*{Extract feature with two PointNet encoders} & $1024$ ($\mathcal{F}^h$) \\ 
 & & $1024$ ($\mathcal{F}^o_g$) \\ 
 & & $3000\times64$ ($\mathcal{F}^o_l$) \\\shline
 \multicolumn{3}{c}{\textbf{Feature fusion}} \\ \hline
 \multirow{2}*{2} & \multirow{2}*{\shortstack{ \textit{concat}(\textit{add}($\mathcal{F}^h$, $\mathcal{F}^o_g$).\textit{repeat}($3000$), \\
 $\mathcal{F}^o_l$)}} & \multirow{2}*{$3000\times1088$} \\ \\\shline
 \multicolumn{3}{c}{\textbf{Contact map regression}} \\ \hline
 \multirow{2}*{3} & \multirow{2}*{\shortstack{1-D convolutions\\ (1088, 512, 256, 128, 1, sigmoid)}} & \multirow{2}*{$3000\times 1$} \\ \\\shline
\end{tabular}
\caption{ContactNet architecture.}
\vspace{-0.15in}
\label{tab: contactnet}
\end{table}

Table~\ref{tab: contactnet} shows the architecture of ContactNet, which takes in both hand-object point cloud to regress the object contact map. In the network, we use both of the object global and local features, $\mathcal{F}^o_g$ and $\mathcal{F}^o_l$, where the local features are used to maintain the point correspondence.

\section*{Appendix B: Details of Experiments and Evaluation}

\subsection*{B.1. Datasets}
We follow~\cite{karunratanakul2020grasping} to use HO-3D and FPHA datasets for evaluating the generalization ability of the proposed method. For FPHA dataset, the ground-truth hand mesh are fitted on the provided hand joints. We follow~\cite{karunratanakul2020grasping} to exclude the huge objects (especially milk bottle) in the FPHA dataset.

\subsection*{B.2. Evaluation Metrics}
\begin{figure}
\centering
\includegraphics[width=8cm]{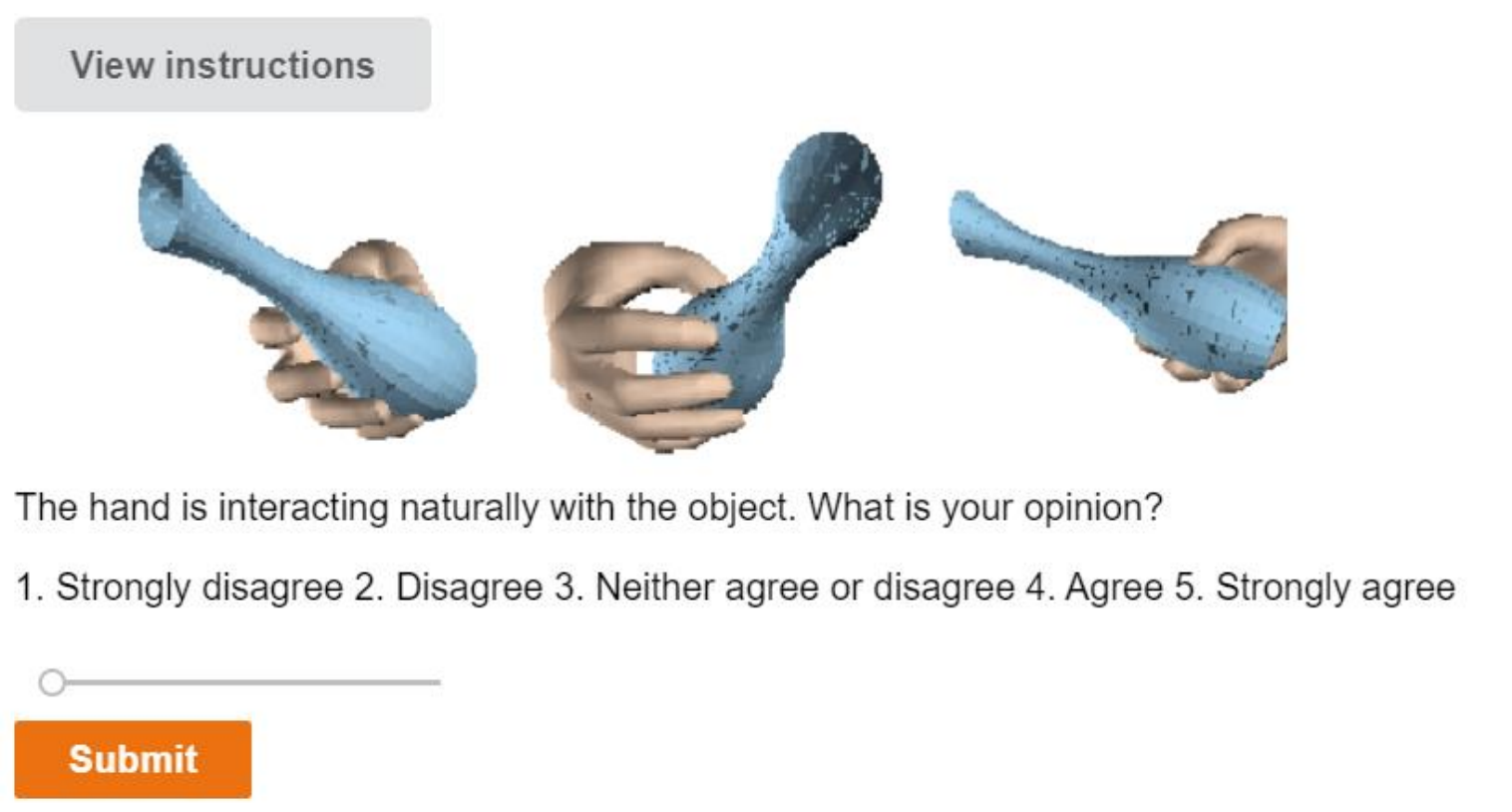}
\vspace{-0.1in}
\caption{AMT online evaluation layout.}
\label{fig:AMT}
\end{figure}

\textbf{Perceptual score.} The perceptual score is evaluated with Amazon Mechanical Turk following~\cite{karunratanakul2020grasping}, the layout is shown in Fig.~\ref{fig:AMT}. We show 3 views of each sample. The rating score ranges from 1 to 5. Every sample is rated by 3 workers.

\textbf{Penetration.}
The penetration is to measure the collision between the hand and the object. We report the 
maximum penetration depth and penetration volume following~\cite{hasson2019learning}. The former is calculated as the largest distance from the penetrating vertices of hand mesh to the closed object surface. And the latter is the volume of the intersecting voxels between the hand and object meshes. To compute this metric, we first voxelize both hand and object mesh using the voxel size of $0.5\ cm$, and then compute the number of intersecting voxels.
The result is computed by the voxel volume times the number of intersecting voxels.

\textbf{Reconstruction Error.} 
We \textbf{do not} use hand reconstruction error (mesh reconstruction error on hand mesh, or kinematics error on hand joints) as a metric for evaluating the quality of generated grasps. Because grasp generation has multiple solutions, a good and reasonable grasp can be far away from the GT (Note that only one GT grasp is provided for each sample in datasets we used). Thus it does not make sense in our case to measure the reconstruction error with only one GT.

\subsection*{B.3. Experiments}
We introduce more experiments details and results in this section, including details of GraspCVAE training targets and Test-time Adaptation.

\subsubsection*{B.3.1. GraspCVAE Training Targets}
\quad Table~\ref{sup: loss_compare} shows the performance of the GraspCVAE trained by losses we proposed and losses from~\cite{taheri2020grab}, which are tested on the Obman test set. Our training targets performs significantly better.

\subsubsection*{B.3.2. Test-time Adaptation}

\quad \textbf{Details of TTA} During TTA, each test sample is adapted in a self-supervised manner for $10$ iterations. For each iteration, the single test object is augmented into a batch which includes $32$ samples, where the augmentation is random translation in $[-5,5]\ cm$. Due to the reason that the augmentation is supposed to maintain the geometry feature of the object, other augmentation methods, e.g. scaling and rotation, are harmful.

\begin{table}[t]
\vspace{-0.1in}
\tiny
\tablestyle{5pt}{1.05}
\begin{tabular}{l|cccc}
 & Penetr Vol. $\downarrow$ & Stability $\downarrow$ & Percep Score $\uparrow$ & Contact ($\%$) $\uparrow$ \\ \shline
 ~\cite{taheri2020grab} & 8.41 & 1.66 & 2.97 & 98.25\\
 Ours & \textbf{5.12} & \textbf{1.52} & \textbf{3.54} & \textbf{99.97}
\end{tabular}
\caption{Performance of the GraspCVAE trained by losses we proposed and losses from~\cite{taheri2020grab} on Obman dataset.}
\label{sup: loss_compare}
\vspace{-0.15in}
\end{table}

\textbf{Details of Different TTA Paradigms}
In the Sec.~4.5.3 of the paper, we compare different TTA paradigms. And we give more details here.
\begin{itemize}
    \item TTA-optm (offline): In this method, we only optimize the 45-D hand joints axis-angle rotation tensor, rather than the 61-D full hand pose parameters as in other learning-based TTA. We observe that optimizing the 61-D full hand pose is not stable, and the results can even become worse.
    \item TTA-noise (offline): In this method, when we train the ContactNet, we injecting random noise on the input 45-D hand joint rotation tensor. The model is denoted as ContactNet-noise. The reconstruction error of ContactNet-noise is \textbf{0.109}, higher than the original \textbf{0.090} without injecting noise (Table~\ref{tab: generalization} in paper). The increased error demonstrate that injecting noise is harmful for learning contact maps, and implies that the network cannot learn to "corret" the noise. It is also the reason for the worse results of TTA-noise compared with original TTA.
    \item TTA-online: The HO-3D and FPHA datasets are video datasets, and the TTA-online is performed on the video clips. Because the object pose changes smoothly in the video frames, it provides the chance for the network to fit the test distribution continuously better. Besides, the TTA-online also demonstrate that the model after TTA \textbf{does not overfit} to the single test sample, because it can continually generates grasps of the following incoming test samples without re-initializing the network parameters.
\end{itemize}

\section*{Appendix C: Additional Results}
\label{sec: appendix_c}

\begin{figure*}[t]
\centering
\includegraphics[width=0.85\linewidth]{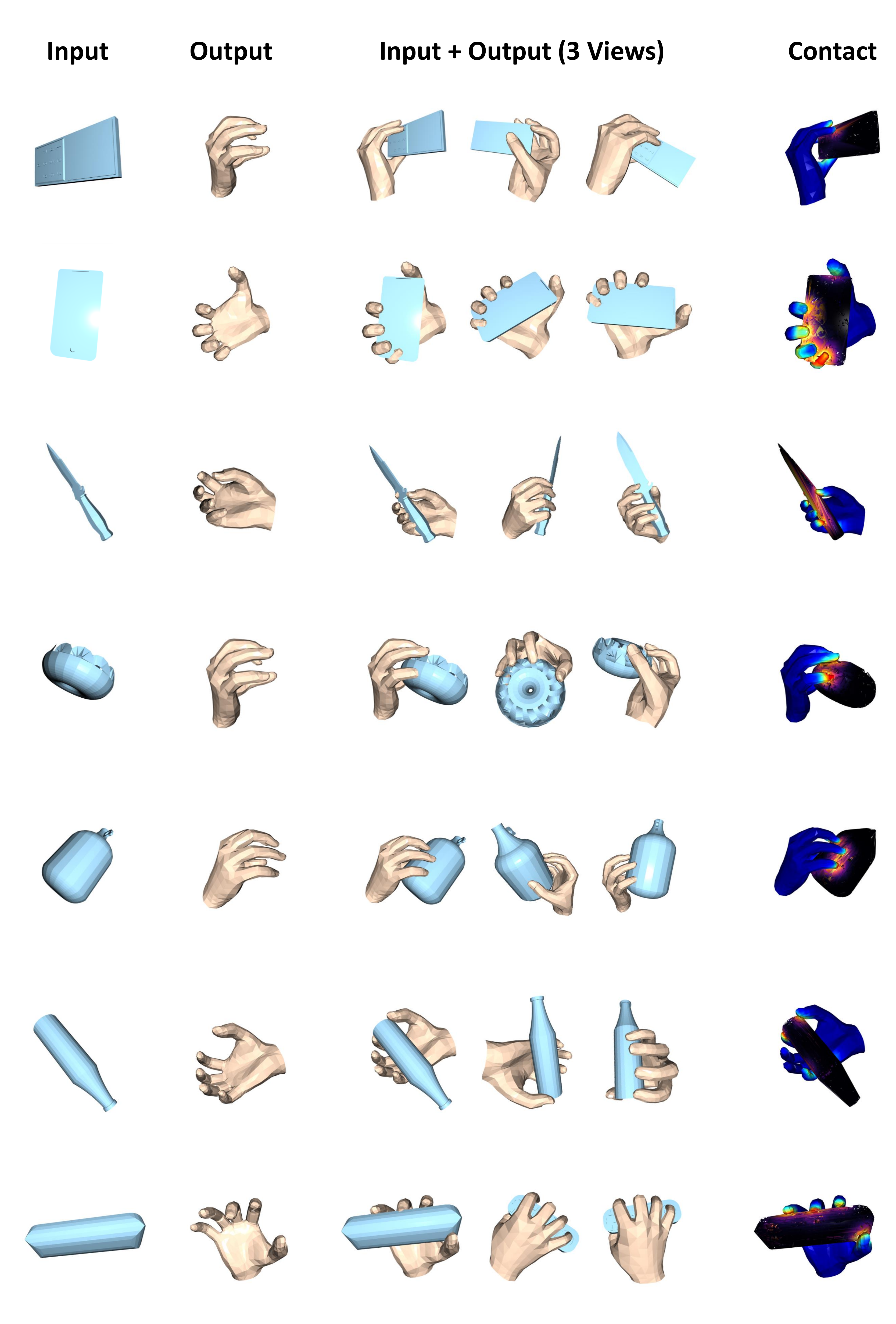}
\vspace{-0.2in}
\caption{Results of generated grasps given \textcolor{obmanblue}{in-domain} objects. Every result is shown in a row with input object, output hand mesh, both input and output in 3 views and in contact.}
\label{fig: sup_results_indomain}
\end{figure*}

\begin{figure*}[t]
\centering
\includegraphics[width=0.85\linewidth]{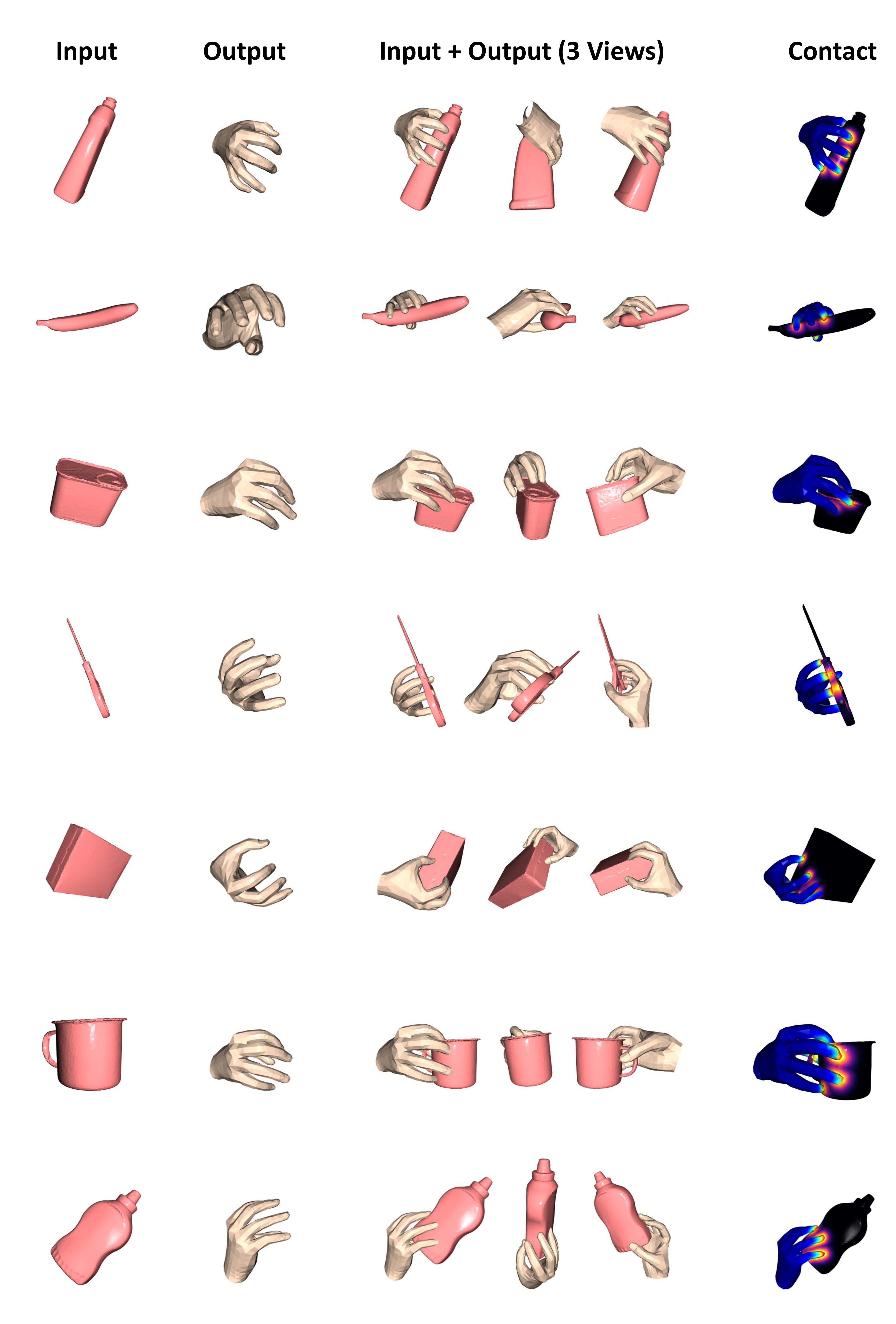}
\vspace{-0.2in}
\caption{Results of generated grasps given \textcolor{hored}{out-of-domain}. Every result is shown in a row with input object, output hand mesh, both input and output in 3 views and in contact.}
\label{fig: sup_results_outofdomain}
\end{figure*}

\textbf{More visualization} are shown in Fig.~\ref{fig: sup_results_indomain} for \textcolor{obmanblue}{in-domain} Obman test set objects, and Fig.~\ref{fig: sup_results_outofdomain} for \textcolor{hored}{out-of-domain} HO-3D objects. Each result is shown in a row. All results are chosen randomly.

\begin{figure*}[t]
\centering
\includegraphics[width=0.95\linewidth]{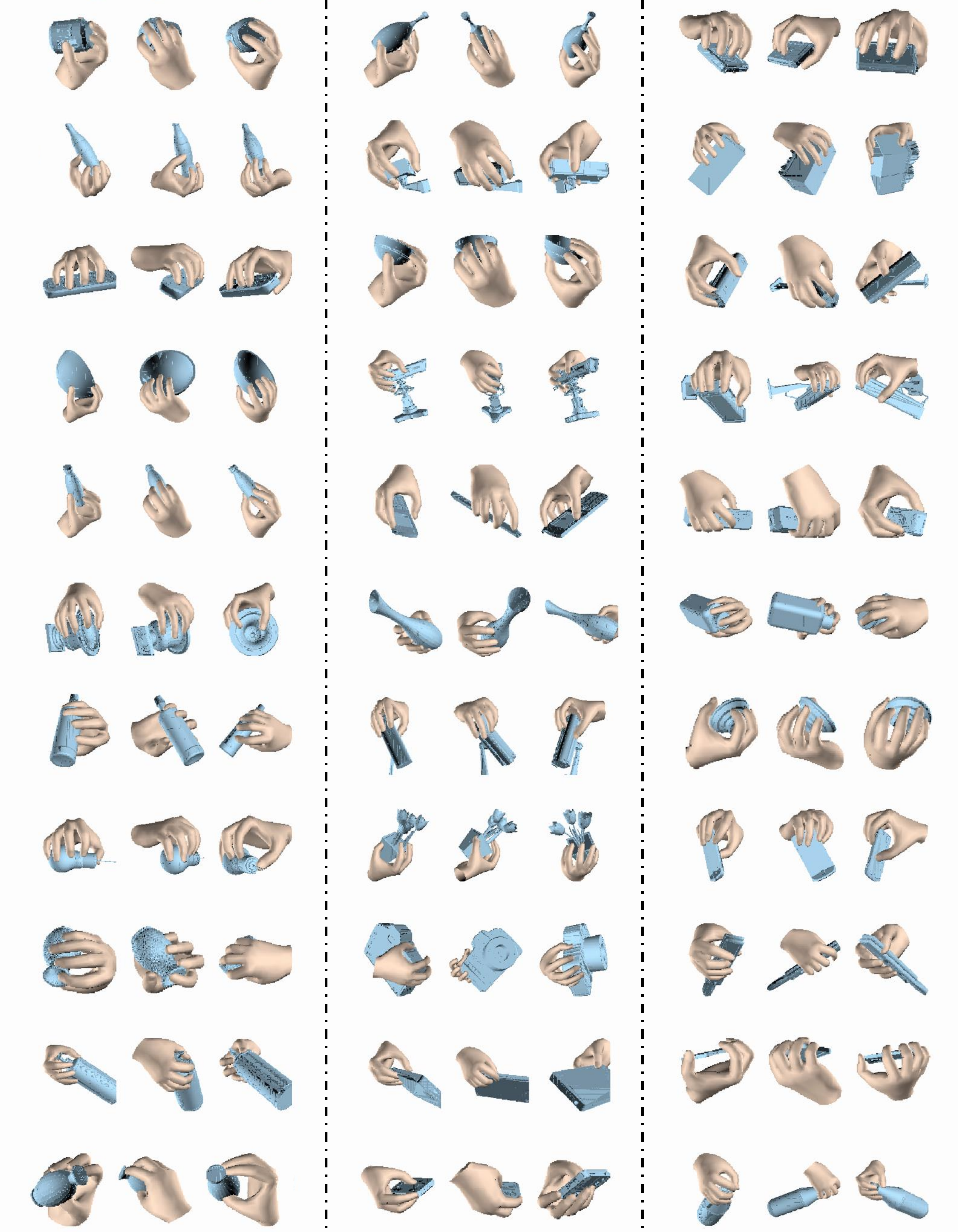}
\caption{Generated grasps given \textcolor{obmanblue}{in-domain} Obman test objects. Each results is shown in 3 views. All results are chosen randomly.}
\label{fig: sup_obman}
\end{figure*}

\begin{figure*}[t]
\centering
\includegraphics[width=0.95\linewidth]{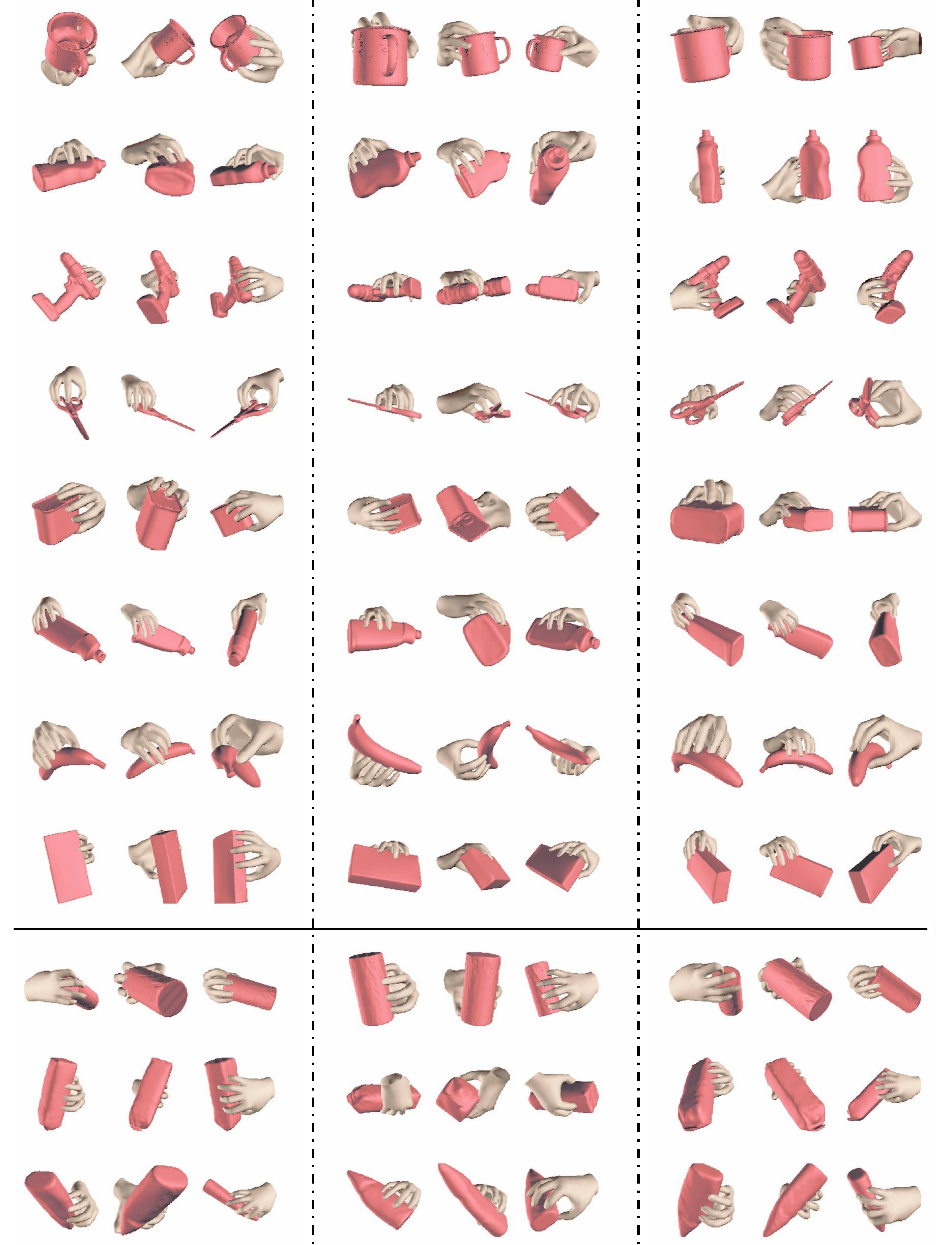}
\vspace{-0.1in}
\caption{Generated grasps on \textcolor{hored}{out-of-domain} HO-3D and FPHA objects. 8 out of 10 objects of HO-3D dataset and all 3 objects of FPHA dataset are visualized. We include 3 results of each object in each row. Each result is shown in 3 views. All results are chosen randomly.}
\label{fig: sup_hoed_fpha}
\end{figure*}

\textbf{More results} are shown in Fig.~\ref{fig: sup_obman} for \textcolor{obmanblue}{in-domain} Obman test set objects, and Fig.~\ref{fig: sup_hoed_fpha} for \textcolor{hored}{out-of-domain} HO-3D and FPHA objects. We show 4 examples in each row, and each result is shown with 3 views. All results are chosen randomly.

\end{document}